\documentclass[twoside]{article}
\usepackage{aistats2019}

\usepackage[round,sort,comma]{natbib} %

\usepackage[utf8]{inputenc} %
\usepackage{url}            %
\usepackage{booktabs}       %
\usepackage{amsfonts}       %
\usepackage{nicefrac}       %
\usepackage{microtype}      %

\usepackage{mathtools}
\usepackage{bm}
\usepackage{dsfont}
\usepackage{color}
\usepackage{graphicx} %
\usepackage{subcaption}
\graphicspath{{final_figures/}, }
\usepackage{enumitem}
\usepackage{balance}

\newcommand{\mysubsection}[1]{\textbf{#1}}
\newcommand{\evalsubsection}[1]{\textbf{#1}:}

\newcommand{\tdist}{p^\ast\!} %\breve{p}}
\newcommand{\tdistx}{\tdist(\vx)}

\newcommand{\truepost}{p_{\theta}(\vz|\vx)}
\newcommand{\zdist}{p(\vz)}
\newcommand{\modeldist}{p_{\vtheta}}

\newcommand{\modelrdistx}{p^R_{\vtheta(\vx)}}
\newcommand{\modeldistx}{p_{\vtheta}(\vx)}

\newcommand{\vardistz}{q_{\eta}(\vz|\vx)}
\newcommand{\qmarg}{q_{\eta}(\vz)}
\newcommand{\vardistzn}{q_{\eta}(\vz|\vx_{n})}

\newcommand\cut[1]{}

\newcommand{\Reals}{\mathbb{R}}

\newcommand{\squishlist}{
   \begin{list}{$\bullet$}
    { \setlength{\itemsep}{0pt}      \setlength{\parsep}{3pt}
      \setlength{\topsep}{3pt}       \setlength{\partopsep}{0pt}
      \setlength{\leftmargin}{1.5em} \setlength{\labelwidth}{1em}
      \setlength{\labelsep}{0.5em} } }

\newcommand{\squishlisttwo}{
   \begin{list}{$\bullet$}
    { \setlength{\itemsep}{0pt}    \setlength{\parsep}{0pt}
      \setlength{\topsep}{0pt}     \setlength{\partopsep}{0pt}
      \setlength{\leftmargin}{2em} \setlength{\labelwidth}{1.5em}
      \setlength{\labelsep}{0.5em} } }

\newcommand{\squishend}{
    \end{list}  }

\newcommand{\KLpq}[2]{\textrm{KL}\!\left[{#1}||{#2}\right]}

\newcommand{\myvec}[1]{\mathbf{#1}}
\newcommand{\myvecsym}[1]{\boldsymbol{#1}}

\newcommand{\veta}{\myvecsym{\eta}}

\newcommand{\vomega}{\myvecsym{\omega}}

\newcommand{\vphi}{\myvecsym{\phi}}

\newcommand{\vtheta}{\myvecsym{\theta}}

\newcommand{\vb}{\myvec{b}}

\newcommand{\vt}{\myvec{t}}
\newcommand{\vu}{\myvec{u}}

\newcommand{\vx}{\myvec{x}}
\newcommand{\vxhat}{\hat{\vx}}

\newcommand{\vy}{\myvec{y}}

\newcommand{\vz}{\myvec{z}}
\newcommand{\vzhat}{\hat{\vz}}

\newcommand{\E}{\mathbb{E}}

\newcommand{\be}{\begin{equation}}
\newcommand{\ee}{\end{equation}}
\newcommand{\bea}{\begin{eqnarray}}
\newcommand{\eea}{\end{eqnarray}}
\newcommand{\beaa}{\begin{eqnarray*}}
\newcommand{\eeaa}{\end{eqnarray*}}

\DeclareMathAlphabet{\mathpzc}{OT1}{pzc}{m}{n}

\newcommand{\disc}{\mathcal{D}_{\vphi}}
\newcommand{\discx}{\mathcal{D}_{\vphi}(\vx)}
\newcommand{\codedisc}{\mathcal{C}_{\vomega}}
\newcommand{\codediscz}{\codedisc(\vz)}

\newcommand{\generator}{\mathcal{G}_{\vtheta}}

\usepackage{afterpage}
\usepackage[dvipsnames]{xcolor}
\usepackage{algorithm}
\usepackage{algcompatible}
\usepackage{algpseudocode}
\algrenewcommand\algorithmicindent{0.5em} %
\newcommand{\algcomment}[1]{\Comment{\textit{#1}}}
\algnewcommand{\LineComment}[1]{\(\triangleright\) \textit{#1}}

\newcommand{\addhspace}{\quad}
\newcommand{\addhspacesmall}{\hspace{1pt}}

\newcommand{\ourgan}{VGH++}
\newcommand{\reconourgan}{VGH}
\newcommand{\colormnist}{ColorMNIST}

\newcommand{\specialcell}[2][c]{%
  \begin{tabular}[#1]{@{}c@{}}#2\end{tabular}}

\newtheorem{theorem}{Theorem}

\newcommand{\NA}{---}

\definecolor{mydarkblue}{rgb}{0,0.08,0.45}
\usepackage[colorlinks,citecolor=mydarkblue,urlcolor=mydarkblue,linkcolor=mydarkblue,filecolor=mydarkblue]{hyperref}
\newcommand{\ourtitle}{Distribution Matching in Variational Inference}

\begin{document}

\twocolumn[

\aistatstitle{\ourtitle}
\aistatsauthor{Mihaela Rosca \quad Balaji Lakshminarayanan \quad Shakir Mohamed}
\aistatsaddress{\{mihaelacr,balajiln,shakir\}@google.com}
\vspace{-5mm}
\aistatsaddress{DeepMind}

]

\begin{abstract}
With the increasingly widespread deployment of generative models, there is a mounting need for a
deeper understanding of their behaviors and limitations. In this paper, we expose
the limitations of Variational Autoencoders (VAEs), which consistently fail to
learn marginal distributions in both latent and visible spaces. We show this to be a consequence of learning by matching conditional distributions, and the limitations of explicit
model and posterior distributions. It is popular to consider Generative
Adversarial Networks (GANs) as a means of overcoming these limitations, leading
to hybrids of VAEs and GANs. We perform a large-scale evaluation of several
VAE-GAN hybrids and analyze the implications of class probability estimation for
learning distributions. While promising, we conclude that at present, VAE-GAN
hybrids have limited applicability: they are harder to scale, evaluate, and use
for inference compared to VAEs; and they do not improve over the generation
quality of GANs.
\end{abstract}

\section{INTRODUCTION}
This paper focuses on the challenges of training Variational Autoencoders (VAEs) \citep{aevb, dlgm}: the
struggles of matching the marginal latent posterior distribution with the prior, the difficulties faced in
explicitly specifying latent posteriors, and the lack of likelihoods that capture the semantic similarity of
data. To overcome the limitations of VAEs, it has become natural to consider borrowing the
strengths of another popular type of generative algorithm, Generative Adversarial networks
(GANs) \citep{gan}, resulting in a fusion of VAEs and GANs \citep{adversarialvb, veegan, adversarialae,
aepixels, alphagan, pu2017adversarial}. What are the limitations of VAEs at present? How can
GANs help? Do VAE-GAN hybrids address the limitations currently being experienced? The aim of this
paper is to offer an answer to these questions.

Generative models currently have a wide range of applications in dimensionality
reduction, denoising, reinforcement learning, few-shot learning, data-simulation and emulation,
semi-supervised learning and in-painting \citep{pca, mathieu2016disentangling, darla,
rezende2016one,kingma2014semi,improvedgan, ppgn}, and they continue to be deployed in new
domains, from drug-discovery to high-energy physics. It thus becomes vital
for us to explore their strengths and shortcomings, and deepen our understanding of the performance and behavior of generative models.

We  will contrast conditional distribution matching in VAEs with marginal distribution matching in
VAE-GAN hybrids. We compare the
performance of explicit distributions in VAEs with the use of implicit distributions in VAE-GAN hybrids.
And we measure the effect of distributional assumptions on what is learned.  We will make the following
contributions:
\vspace{-3mm}
\begin{itemize}[leftmargin=*,noitemsep]
  \item We show that VAEs fail to match marginal distributions in both latent and visible space and that powerful explicit model distributions and powerful explicit posteriors do not improve marginal distribution matching in VAEs. This is prevalent across datasets, models and latent dimensionality.
  \item We systematically evaluate existing and new VAE-GAN hybrids as promising avenues to improve variational inference, but show that since they use classifier probabilities to estimate density ratios and learn implicit distributions, they lack an accurate estimate for the likelihood bound that can be used for model evaluation and comparison.
  \item We uncover the effect of marginal distribution matching in latent space on latent representations
  and learned posterior distributions, and show that VAE-GAN hybrids are harder to scale to higher latent
  dimensions than VAEs.
\end{itemize}
These contributions will lead us to conclude that while VAE-GAN hybrids allow for posterior inference in
implicit generative models, \textit{at present}, VAE-GAN hybrids have limited applicability: compared to
VAEs, their use of classifier probabilities makes them harder to scale, evaluate, and use for inference;
compared to GANs, they do not improve sample generation quality.

\section{CHALLENGES WITH VAES}
\label{sect:problem_statement}
One goal of generative models is to match the unknown distribution of data $\tdistx$ to the distribution
learned by a model. In latent variables models, $\modeldist(\vx)$ is defined via a latent variable $\vz$
and the hierarchical model:
\begin{align}
\modeldist(\vx)  = \int \modeldist(\vx|\vz)\zdist d\vz.
\label{eq:marginal_x}
\end{align}
The integral in Equation~\ref{eq:marginal_x} for computing $\modeldist(\vx)$ is, in general, intractable,
making it hard maximize the marginal likelihood of the model under the data,
$\mathbb{E}_{\tdistx}[\log \modeldist(\vx)]$. To overcome this intractability, variational inference
introduces a
tractable lower bound on $\modeldist(\vx)$ via a variational distribution $\vardistz$:
\begin{align}
& \log \modeldist(\vx) - \KLpq{\vardistz}{\truepost} \nonumber \\
& = \E_{\vardistz} [ \log \modeldist(\vx|\vz) ] - \KLpq{\vardistz}{\zdist} \label{eq:all_model} \\
& \!\!\!\!\log \modeldist(\vx) \!\!\geq\! \underbrace{\E_{\vardistz} [ \log \modeldist(\vx|\vz) ]}_{\text{likelihood
term}} \!-\! \underbrace{\KLpq{\vardistz}{\zdist}}_{\text {KL penalty}} \label{eq:free_energy}
\end{align}
The training objective is to learn the model parameters $\theta$ of the distribution $\modeldist(\vx|\vz)$ and the variational parameters $\veta$ of $\vardistz$ to maximize the evidence lower bound (ELBO):
\begin{align}
 \mathbb{E}_{\tdistx} \left[\mathbb{E}_{\vardistz} [ \log \modeldist(\vx|\vz) ] - \KLpq{\vardistz}{\zdist}\right]
 \label{eq:vae_objective}
\end{align}
Distributions $\vardistz$ and $\modeldist(\vx|\vz)$ can be parametrized using neural networks and
trained jointly using stochastic gradient descent as in Variational Autoencoders  \citep{aevb, dlgm}. In
this setting, we refer to $\vardistz$ as the encoder distribution, and to $\modeldist(\vx|\vz)$ as the
decoder distribution.

The ideal learning scenario does \textit{marginal distribution matching}, in visible space: matching
$\modeldist(\vx)$ to the true data distribution $\tdistx$. Variational inference, using a lower bound
approximation, achieves this indirectly using  \textit{conditional distribution matching} in
latent space: minimizing $\KLpq{\vardistz}{\truepost}$ in Equation~\ref{eq:all_model} and searching
for a tight variational bound.

Marginal distribution matching in \textit{visible} space, and conditional distribution matching in
\textit{latent} space, are
tightly related through marginal distribution matching in latent space
By doing a ``surgery on the ELBO'', \citet{elbo_surgery} showed how variational inference methods
minimize $\KLpq{\qmarg}{\zdist}$ when maximizing the ELBO, since:
\begin{align}  \label{eq:mi}
&  \mathbb{E}_{\tdistx} \KLpq{\vardistz}{\zdist} = \\
& \KLpq{\qmarg}{\zdist} + \int \vardistz \tdistx \log \frac{\vardistz}{q(\vz)}  d\vz d\vx \nonumber
\end{align}
The integral in \eqref{eq:mi} is a mutual information and is non-negative. Minimizing
$\mathbb{E}_{\tdistx} \KLpq{\vardistz}{\zdist}$ therefore minimizes a lower bound on
$\KLpq{\qmarg}{\zdist}$, i.e. matching the marginal distributions.

There are two reasons to be interested in marginal distribution matching. Firstly, we can unveil a failure to
match $\qmarg$ to $\zdist$, and $\modeldist(\vx)$ to $\tdistx$ in VAEs. This  failure can be
associated to the use of conditional distribution matching and explicit distributions. Our goal will be to
explore the solutions to overcome these failures provided by marginal distribution matching and implicit
models. Secondly, the marginal divergence $\KLpq{\qmarg}{\zdist}$ can be used as a metric of VAE
performance since it captures simultaneously two important characteristics of performance: the ability of
the learned posterior distribution to match the true posterior, and the ability of the model to learn the
data distribution.

An alternative examination of VAEs as a rate-distortion analysis, achieved by using a coefficient in front of
the KL term in the ELBO (e.g., \citet{betavae}), is the closest to our work.
\citet{alemi2017information} show that simple decoders have a smaller KL term but have high
reconstruction error, whereas complex decoders are better suited for reconstruction. Their investigation
does not consider implicit distributions and alternative approaches to distribution matching, and is
complementary to this effort.

\section{QUANTIFYING VAE CHALLENGES}
\label{sec:marginal_kl}
To motivate our later exploration of marginal distribution matching and implicit distribution in variational
inference, we first unpack the effects of conditional distribution matching and explicit distributions. We
show that VAEs are unable to match the marginal latent posterior $\qmarg$ to the prior $p(\vz)$. This
will result in a failure to learn the data distribution, and manifests in a discrepancy in quality between
samples and reconstructions from the model. To the best of our knowledge, no other extensive study has
been performed showing the prevalence of this issue across data sets, large latent sizes, posterior and
visible distributions, or leveraged this knowledge to generate low posterior-probability VAE samples.
\citet{elbo_surgery} initially showed VAEs with Gaussian posterior distributions, Bernoulli visible
distribution and  a small number of latents trained on binary MNIST do not match $\qmarg$ and $p(\vz)$.

We train VAEs with different posterior and model distributions on \colormnist~\citep{unrolledgan}, CelebA
\citep{celeba} at 64x64 image resolution, and CIFAR-10 \citep{cifar10}. Throughout this section, we
estimate $\qmarg$ via Monte Carlo using $\frac{1}{N} \sum_{n=1}^{N} q(\vz | \vx_{n})$ (pseudocode in
Appendix~\ref{app:marginal_kl_estimation}).

\subsection{Effect of latent posterior distribution}
\label{sec:elbo_surgery_large_datasets}
\begin{figure*}[t]
	\centering
	\begin{subfigure}{.3\textwidth}{
			\centering
			\includegraphics[width=\textwidth]{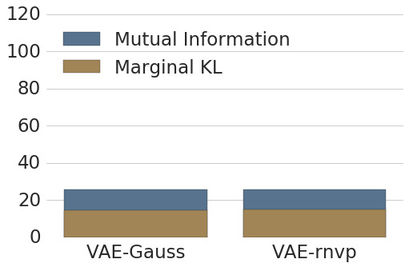}
			\caption{Color-MNIST}
	}\end{subfigure}
	\hspace{3mm}
	\begin{subfigure}{.3\textwidth}{
			\centering
			\includegraphics[width=\textwidth]{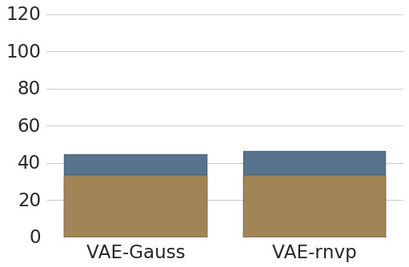}
			\caption{CIFAR-10}
	}\end{subfigure}
	\begin{subfigure}{.3\textwidth}{
			\centering
			\includegraphics[width=\textwidth]{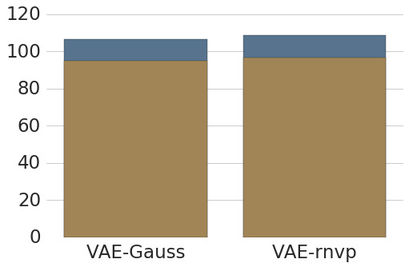}
			\caption{CelebA}
	}\end{subfigure}
	\caption{Marginal KL and mutual information of VAEs with Bernoulli visible distribution and Gaussian
	and RNVP posteriors, adding up to the KL term in the ELBO. $\KLpq{\qmarg}{\zdist}$ reflects the
	complexity of the data: all models use 160 latents, but complex datasets exhibit higher values.}
	\label{fig:marginal_kl_failure}
\end{figure*}
Powerful explicit posteriors given by Real Non Volume Preserving (RNVP) normalizing
flows~\citep{dinh2016density} do not improve marginal distribution matching in VAEs over simple
diagonal Gaussian posteriors. This result, shown in Figure~\ref{fig:marginal_kl_failure} which reports
$\KLpq{\qmarg}{\zdist}$ for trained VAEs, is surprising given that RNVP posteriors are universal
distribution approximators, and have the capacity to fit complex posteriors $\truepost$ and perhaps
suggests that posterior distributions are not the biggest bottleneck in VAE training.

\subsection{Effect of the visible distribution}
\label{sec:pixel_dist}
The choice of visible distribution $\modeldist(\vx|\vz)$ also affects distribution matching in latent space.
We trained the same VAE architecture using Bernoulli and Quantized Normal visible distributions on
CIFAR-10. A Quantized Normal distribution is a normal distribution with uniform noise $\vu \in [0, 1]$
added to discrete pixel data \citep{noteevaluation}. Since a Bernoulli distribution provides the same
learning signal as a QuantizedNormal with unit variance (formal justification in
Appendix~\ref{app:qn_bernoulli}), the QuantizedNormal distribution has a higher modeling capacity,
leading to better data reconstructions. However, this does not result an increased sample quality: better
reconstructions force $\vardistz$ to encode more information about the data, making the KL between
$\vardistz$ and the data agnostic $\zdist$ higher, thus increasing the gap between sample and
reconstruction quality. Apart from visually assessing the results (see Appendix~\ref{app:qn_bernoulli}),
we can now quantify the different behavior using $\KLpq{\qmarg}{\zdist}$, which is 44.7 for a Bernoulli
model, and 256.7 for the same model with a QuantizedNormal distribution. We saw similar results when
comparing Categorical and Bernoulli distributions.

\subsection{Low probability posterior  samples}
\begin{figure*}[ht]
\centering
\begin{subfigure}{.32\textwidth}{
\centering
\includegraphics[width=0.9\textwidth]{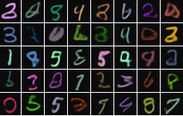} \\
\vspace{0.8mm}
\includegraphics[width=0.9\textwidth]{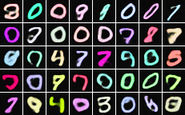}
\caption{Color-MNIST}
}\end{subfigure}
\begin{subfigure}{.32\textwidth}{
\centering
\includegraphics[width=0.9\textwidth]{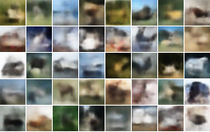} \\
\vspace{0.8mm}
\includegraphics[width=0.9\textwidth]{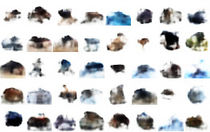}
\caption{CIFAR-10}
}\end{subfigure}
\begin{subfigure}{.32\textwidth}{
\centering
\includegraphics[width=0.9\textwidth]{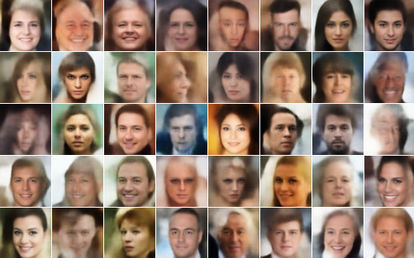} \\
\vspace{0.8mm}
\includegraphics[width=0.9\textwidth]{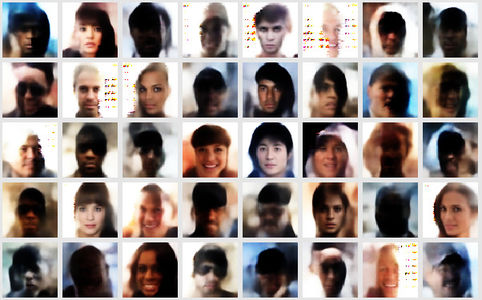}
\caption{CelebA}
}\end{subfigure}
\caption{VAE samples (top) and low posterior VAE samples (bottom). More samples in Appendix~\ref{app:breaking_vaes}.}
\label{fig:bad_samples_all}
\end{figure*}

\begin{figure}[htb!]
	\begin{center}
		\begin{subfigure}{0.55\textwidth}{
				\includegraphics[width=0.4\textwidth]{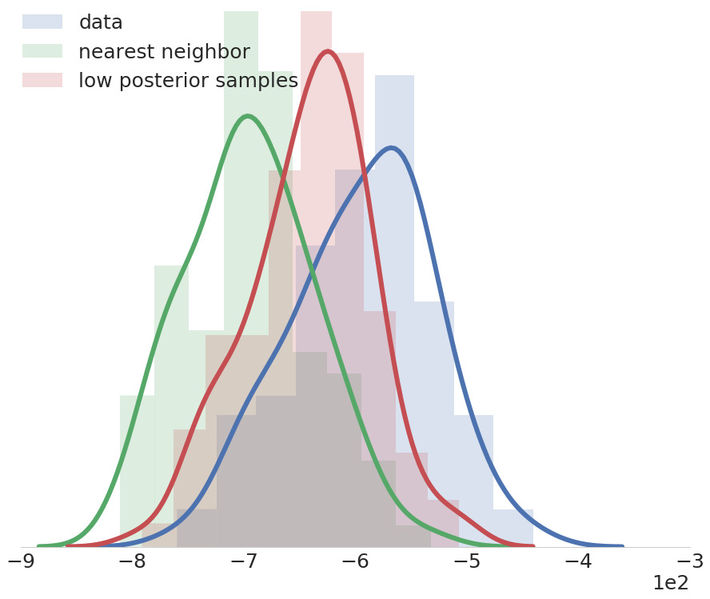}
				\includegraphics[width=0.4\textwidth]{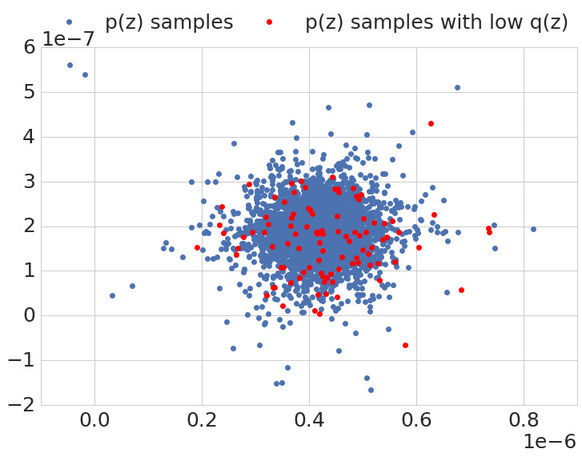}
				\label{fig:cmnist:bad_samples_elbo}
				\caption{ColorMNIST}
		}\end{subfigure}
		\begin{subfigure}{0.55\textwidth}{
				\includegraphics[width=0.4\textwidth]{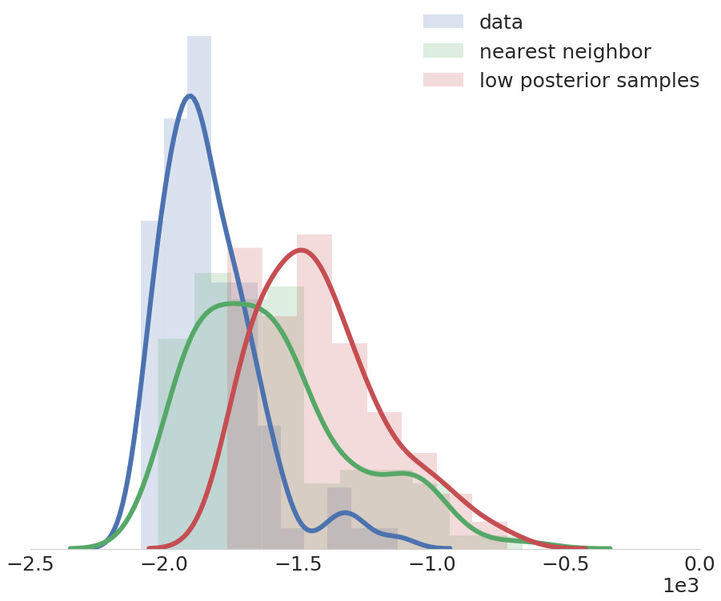} \addhspacesmall
					\includegraphics[width=0.4\textwidth]{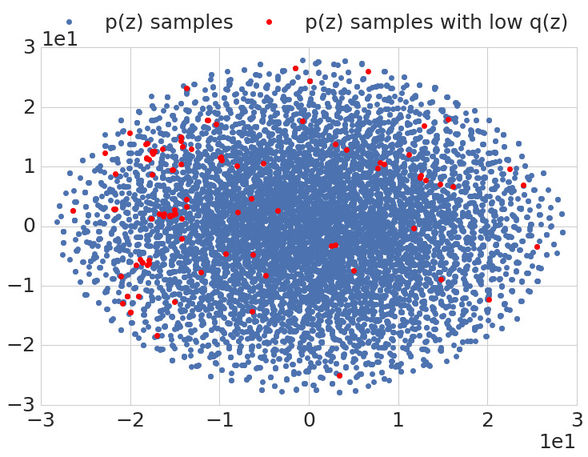} \addhspacesmall
				\caption{CIFAR-10}
				\label{fig:cifar:bad_samples_elbo}
		}\end{subfigure}
		\begin{subfigure}{0.55\textwidth}{
				\includegraphics[width=0.4\textwidth]{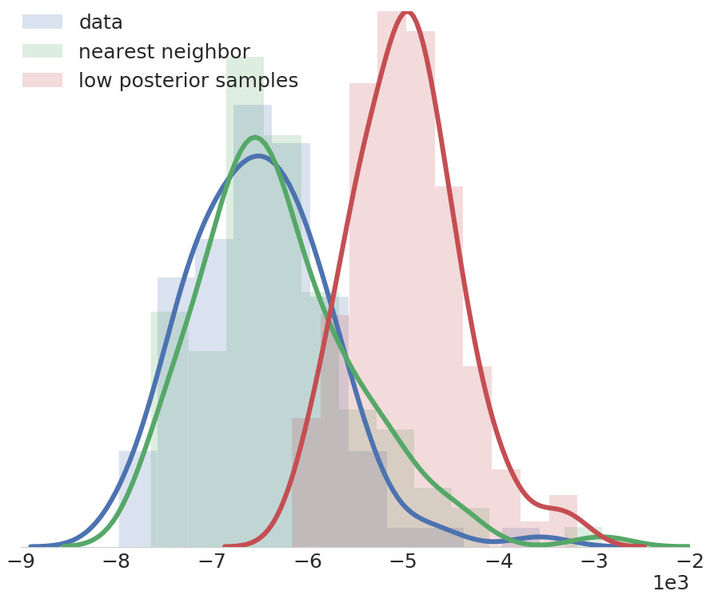} \addhspacesmall
					\includegraphics[width=0.4\textwidth]{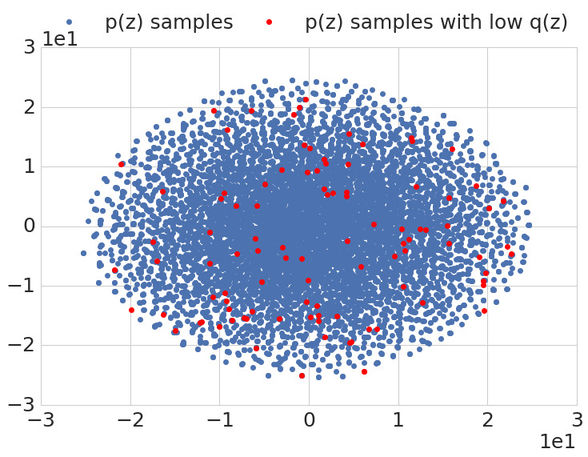} \addhspacesmall
				\caption{CelebA}
				\label{fig:celeba:bad_samples_elbo}
		}\end{subfigure}
		\caption{(Left)Evidence lower bound of uniformly sampled data, nearest neighbors to the
		low posterior samples from the dataset, and the low-probability posterior samples. We also plot the
		result of using a kernel density estimation for each histogram. (Right) 2D visualizations of the latent
		vectors used to create the VAE low posterior
		samples obtained using t-SNE.}
		\label{fig:elbo_bad_samples}.
	\end{center}
\end{figure}

We exploit the failure of VAEs to match marginal distributions in latent space to purposefully generate
`bad' samples: samples obtained from prior samples $z$ that have low probability under the marginal
posterior $\qmarg$. Such
bad samples, shown in Figure~\ref{fig:bad_samples_all}, exhibit common characteristics such as having thickened-lines for ColorMNIST and strong white backgrounds for CelebA and CIFAR-10.
We can look at \textit{where} in latent space the samples with low probability under the
marginal are. Figure \ref{fig:elbo_bad_samples} (right) is a t-SNE visualisation which shows that the
latents which generate the low posterior VAE samples in Figure~\ref{fig:bad_samples_all} are scattered throughout the space of the prior and not isolated to any particlar regions, with large pockets of prior space which are not covered by $\qmarg$.

Since these `bad' samples are unlikely under the true distribution, we expect a well trained model to distinguish easily, using the likelihood bound, typical images (which come from the data) from atypical samples (generated from regions of the marginal distribution with low-probability). Figure \ref{fig:elbo_bad_samples} (left) shows that on ColorMNIST the model recognizes that the low posterior samples and their nearest neighbors have a lower likelihood than
the data, as expected. But for CIFAR-10 and CelebA the model thinks the low posterior samples are more
likely than the data, despite the samples being quantitatively and qualitatively different (see
Appendix~\ref{app:breaking_vaes} for more samples and metrics).

The latent space spread of low posterior samples, together with the inability of VAEs to learn a likelihood that reflects the data distribution, show a systematic failure of these models to match distributions in latent and visible space.

\section{GANS IN LATENT SPACE}
\vspace{-3mm}
\label{sec:density_ratios_latent_space}
If it is conditional divergence minimization and explicit posteriors that result in a failure of VAEs to match
marginal distributions, then we are strongly motivated to explore other approaches . We turn to another
popular type of algorithm, generative adversarial networks (GANs), and explore \textit{the density ratio
trick}, which they use as a tool for marginal distribution matching and implicit distributions in variational
inference, leading to VAE-GAN hybrids.

\subsection{Distribution matching with density ratios}

The density ratio trick \citep{sugiyama2012density,gan, distinguishability} leverages the power of classifiers in order to estimate density ratios. Under the assumption that we can train a perfect binary classifier $\mathcal{D}$ to associate the label $y=1$ to samples from $p_1(\vx)$ and the label $y=0$ to samples from $p_0(\vx)$, the following holds:
\begin{equation}
\small
\frac{p_1(\vx)}{p_0(\vx)} = \frac {p(y=1|\vx)}{p(y=0|\vx)} = \frac{\mathcal{D}}{1-\mathcal{D}}
\label{eq:dr}
\end{equation}
Conveniently, this approach only requires distribution samples, without the explicit forms of the two
distributions. GANs \citep{gan} use the density ratio trick to directly match the data distribution $\tdistx$
with the marginal model distribution $p_{\theta}(\vx)$. By making use of \emph{implicit latent variable
models} \citep{implicitgen} GANs do not require observation likelihoods - they define $p_{\theta}(\vx)$ by
specifying a deterministic mapping $G: \vz \rightarrow \vx$ used to generate model samples. GANs learn
via an adversarial game, using a discriminator to distinguish between generated samples and real data. In
the original formulation, the training was given by the min-max bi-level optimization with value
function\footnote{In all our experiments we use the alternative non-saturating generator loss
$\E_{\modeldistx}[-\log\discx]$~\citep{gan}.}:\\
\begin{equation*}
 \min_G \max_D \mathbb{E}_{\tdistx}\bigl[\log D(\vx)\bigr] + \E_{p(\vz)}\bigl[\log (1 -D(G(\vz)))\bigr]
 \label{eq:gan_min_max}
\end{equation*}
When the density ratio trick is used for learning - like in GANs - the ratio estimator cannot be trained to
optimality for each intermediate model distribution, since this would be computationally prohibitive. In
practice, the model and the estimator are trained jointly using alternating gradient descent
\citep{gan,many_paths}.
Since KL divergences are an expectation of density ratios, the density ratio trick opens the door to implicit posteriors and marginal distribution matching in latent space for variational inference.

\textbf{Implicit variational posteriors.} The choice of posterior distribution in VAEs is limited by the use of
the choice of $\log \vardistzn$ in $\mathbb{E}_{\tdistx} \left[ \KLpq{\vardistzn}{\zdist} \right]$. Since the
density ratio trick only requires samples to estimate $\KLpq{\vardistzn}{\zdist}$, it avoids restrictions on
$\vardistz$ and opens the door to implicit posteriors. However, replacing each KL divergence in
equation~\ref{eq:vae_objective} with a density ratio estimator is not feasible, as this requires an estimator
per data point. AdversarialVB \citep{adversarialvb} solves this issue by training one discriminator to
estimate all ratios $\vardistz / \zdist$ for $x \sim \tdistx$ by distinguishing between pairs $\left(\vx \sim
\tdistx, \vz \sim \zdist \right)$ and  $\left(\vx \sim \tdistx, \vz  \sim \vardistz\right)$.

\textbf{Marginal distribution matching.} Since conditional distribution matching fails to match marginal
distributions in latent and visible space, directly minimizing $\KLpq{\qmarg}{\zdist}$ provides a compelling
alternative. Adversarial Autoencoders (AAE) \citep{adversarialae} ignore the mutual information term in
Equation~\ref{eq:mi} and minimize $\KLpq{\qmarg}{\zdist}$ using adversarial training. AAEs are connected
to marginal distribution matching not only by the ELBO, but also via Wasserstein distance
\citep{tolstikhin2017wasserstein} (see Appendix~\ref{app:wasserstein_autoencoders}).

\subsection{Effects of density ratios on evaluation, scalability and latent representations}
\begin{figure}[t]
\centering
\includegraphics[width=0.23\textwidth]{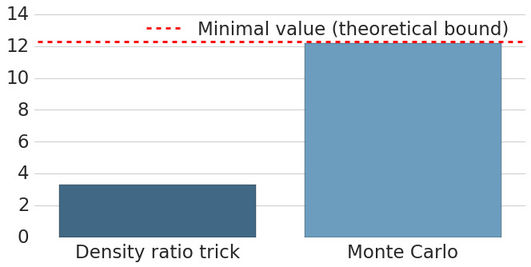}
\includegraphics[width=0.23\textwidth]{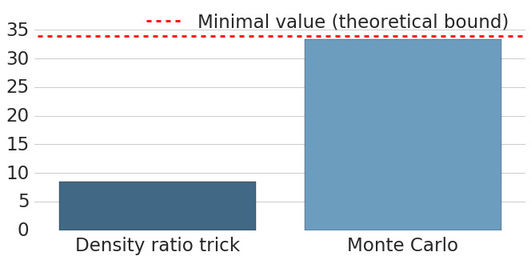}
\caption{The density ratio estimator underestimates $\KLpq{\qmarg}{\zdist}$ for (left) Color-MNIST;
(right) CIFAR.}
\label{fig:density_estimation_marginal_kl_failure}
\vspace{-5mm}
\end{figure}

\begin{figure*}[htb]
	\centering
\begin{subfigure}{.3\textwidth}{
\includegraphics[width=\textwidth]{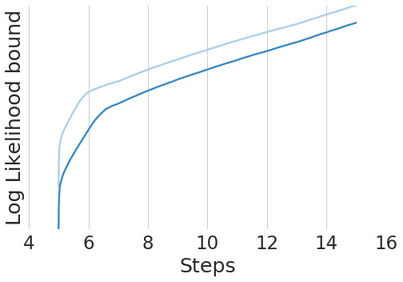}
\caption{VAE}
}\end{subfigure}
\begin{subfigure}{.3\textwidth}{
\includegraphics[width=\textwidth]{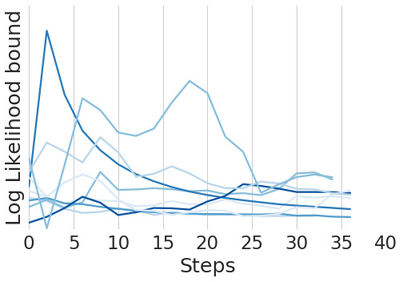}
\caption{AAE}
}\end{subfigure}
\begin{subfigure}{.3\textwidth}{
\includegraphics[width=\textwidth]{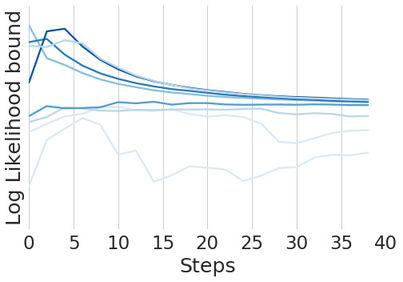}
\caption{Adversarial VB}
}\end{subfigure}
\caption{
   Variational lower bound estimates during training on~\colormnist~across hyperparameters. For accurate estimates, we expect that as model training progresses, the likelihood increases. Similar AdversarialVB results were showed in a maximum likelihood setting, see Figure 18 in~\cite{ivogan}.}
\label{fig:bound_tracking_ae}
\end{figure*}

\textbf{Evaluation}. Using the density ratio trick in variational inference comes at a price: the loss of an estimate for the variational lower bound. Leveraging binary classifiers to estimate KL divergences results in underestimated KL values, even when the discriminator is trained to optimality (see Figure~\ref{fig:density_estimation_marginal_kl_failure}). This issue is exacerbated in practice, as the discriminator is updated online during training. By using density ratios, VAE-GAN hybrids lose a meaningful estimate to report and a quantity to use to assess convergence, as shown in Figure~\ref{fig:bound_tracking_ae}. While VAEs can be compared with other likelihood models, VAE-GAN hybrids can only use data quality metrics for evaluation and lack a metric that can used to track model convergence.

\begin{figure}[ht]
\centering
\includegraphics[width=0.23\textwidth]{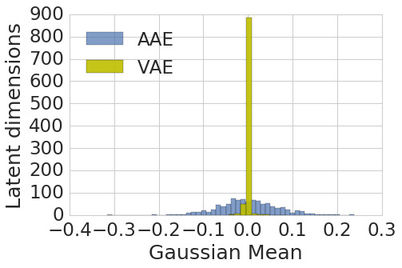}
\includegraphics[width=0.23\textwidth]{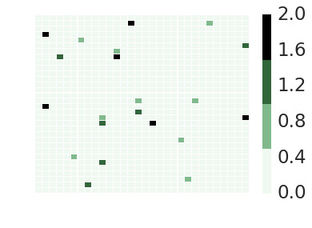}
\caption{The effects of marginal distribution matching on latent representations. (Left) Learned latent
means. (Right) Hinton diagram, showing VAE posterior KLs for dimension of the latent variable reshaped
as a matrix for visualization, and showing that most KL values are low and close to zero.}
\label{fig:rep_learning}
\end{figure}

\textbf{Scalability} is another concern when using the density ratio trick. Using synthetic experiments
where the true KL divergence is known, we show that using classifiers to provide learning signal does not
scale with data dimensions (Appendix~\ref{sec:density_ratio_gradients}). In practice this can be observed
by training VAEs and AAEs with 1000 and 10000 latents. While VAEs learn to ignore extra latents
(Figure~\ref{fig:rep_learning}) and scale with no architectural changes to extra dimensions, AAEs struggle
to model~\colormnist~and completely fail on CelebA even with a bigger discriminator -- see
Figures~\ref{fig:samples_10k} and~\ref{fig:samples_8k_celeba} in Appendix~\ref{app:scaling}.

\textbf{Latent representations} learned via marginal divergence minimization have different properties than
representations learned with VAEs. Figure~\ref{fig:rep_learning} contrasts the posterior Gaussian means
learned using AAEs and VAEs with 1000 latents on~\colormnist: VAEs learn sparse representations, by
using only a few latents to reconstruct the data and ignoring the rest, while
AAEs learn dense representations. This is a consequence of marginal distribution matching: by not having a
cost on the conditional $\vardistz$, AAEs can use any posterior latent distribution but lose the regularizing
effect of $\KLpq{\vardistzn}{\zdist}$ - when learning a Gaussian $\vardistzn$ the variance collapses to 0
and the mean has a much wider range around the prior mean compared to VAEs. This finding suggests that for downstream applications such as semi supervised learning, using VAE representations might have a different effect compared to using representations learned by VAE-GAN hybrids.

\section{GANS IN VISIBLE SPACE}
\label{sect:visble_space}
The benefits and challenges of marginal distribution matching and implicit distributions in variational
inference go beyond matching distributions in latent space. Using \textit{implicit models for the data}
avoids notorious problems with explicit likelihoods, such as blurry samples and reconstructions produced
by common choices such as Gaussian or Laplacian distributions.
To introduce marginal distribution matching in variational inference, we introduce a ratio in the likelihood term of \eqref{eq:vae_objective}:
 \begin{align}
 &\E_{\tdistx} \E_{\vardistz} \left[ \log \modeldist(\vx|\vz)\right] \nonumber \\
& = \E_{\tdistx} \E_{\vardistz} \left[ \log \tfrac{\modeldist(\vx|\vz)}{\tdistx}\right]  +  \mathbb{H}[\tdistx]
 \label{eq:synthetic_lik}
 \end{align}
$\mathbb{H}[\tdistx]$ denotes the entropy of $\tdistx$ and can be ignored for optimization purposes. This approach is a form of synthetic likelihood \citep{wood2010statistical,dutta2016likelihood}.

Equation~\eqref{eq:synthetic_lik} shows the challenges of replacing the likelihood term with a density ratio:
\vspace{-0.5em}
\begin{enumerate}[wide=0.1pt]
\setlength\itemsep{-0.34em}
    \item \label{itm:one_vs_many} Naively using the density ratio trick requires an infinite number of discriminators, one for each $\modeldist(\vx|\vz)$ with $\vz \sim q(\vz|\vx_{n})$.
    \item \label{itm:gradients} The model $\modeldist(\vx|\vz)$ no longer receives gradients, since the parametric estimator replacing $\frac{\modeldist(\vx|\vz)}{\tdistx}$ is evaluated at data points, as given by the expectations in Equation~\ref{eq:synthetic_lik}. The model has been absorbed by the ratio estimator - by providing samples to it.
\end{enumerate}
To avoid both problems and continue analyzing the effects of marginal distribution matching in variational inference, we leverage GANs to train the generative model. A previously proposed solution explores joint distribution matching~\citep{veegan,ali,bigan,li2017alice}.

\subsection{Marginal distribution matching}
\label{sec:synth_like}
To bypass Equation~\eqref{eq:synthetic_lik}, instead of using conditional distribution matching done when maximizing the ELBO, we use a ratio estimator to distinguish between $\tdistx$ and the marginal reconstruction distribution $\modelrdistx = \int \modeldist(\vx|\vz) \qmarg dz$. The loss function for the ratio estimator $\discx$ becomes:
\vspace{-1mm}
\begin{equation}
\mathbb{E}_{\tdistx}\bigl[- \log \discx\bigr] + \E_{\modelrdistx}
\bigl[-\log (1 -\discx)\bigr] \label{eq:gan_d_recon_ours}
 \end{equation}

To produce realistic reconstructions which fool the discriminator, the encoder and decoder minimize an adversarial loss, such as $ - E_{\modelrdistx}\log (1 -\discx)$.
This approach to marginal distribution matching in visible space can be used in conjunction with either conditional or marginal matching in latent space and lends itself to training both implicit and explicit models.

When learning explicit models, marginal distribution matching and conditional distribution matching as done in variational inference can be combined. From the variational inference view the adversarial loss acts as a regularizer, steering reconstructions towards an area of the space that makes them realistic enough to fool the discriminator. From the adversarial perspective, the reconstruction loss can be viewed as a regularizer which avoids mode collapse, a notorious issue with GANs \cite{veegan,wgan}.

All VAE-GAN hybrids discussed so far rely on the KL term to match distributions in latent space. Adversarial training in visible space gives us another way to match $\qmarg$ and $\zdist$, by matching the data distribution and the model distribution, via a discriminator which minimizes:
\begin{align}
\hspace{-3mm}
& 2 \, \mathbb{E}_{\tdistx}\bigl[- \log \discx\bigr] + \E_{\modelrdistx}\bigl[-\log (1 -\discx)\bigr] \nonumber \\
& + \E_{\modeldistx}\bigl[-\log (1 -\discx)\bigr] \label{eq:gan_d_ours}
\end{align}
The decoder now receives an adversarial loss both for reconstructions and samples, learning how to produce compelling looking samples from early on in training, like GANs. Unlike GANs, this model is able to do inference, by learning $\vardistz$. We will later show that this loss helps improve sample quality compared to using an adversarial loss only on reconstructions, as described in equation~\ref{eq:gan_d_recon_ours}.

\subsection{Joint variants}

 Matching \textit{joint} distributions matches both visible and latent space distributions. Minimizing divergences in the joint space leads to solutions for lack of gradients and challenges of density ratio estimation of Equation~\ref{eq:synthetic_lik} while staying completely in the variational inference framework, by changing the goal of the model and using a different variational bound. VEEGAN \cite{veegan} changes the model objective from matching the marginals $\tdistx$ and $\modeldist(\vx)$ in data space to matching marginal distributions in latent space. Using a reconstructor network to learn a posterior distribution $p_{\gamma}(\vz|\vx)$, the goal of VEEGAN is to match $p_{\gamma} (\vz)$ to the prior $p(\vz)$. Because $p_{\gamma} (\vz)$ is intractable, $\modeldist(\vx|\vz)$ is introduced as a variational distribution using the lower bound:
\begingroup
\setlength\abovedisplayskip{0pt}
\setlength{\belowdisplayskip}{0pt}
\begin{equation*}
\int p_{\gamma}(\vz|\vx) \tdistx d\vx \leq \KLpq{\modeldist(\vx|\vz)p(\vz)}{p_{\gamma}(\vz|\vx)\tdistx} + C
\end{equation*}
\endgroup
where $C$ is a constant. The model is trained to minimize the negative of the lower bound and a reconstruction loss in latent space. To estimate $\KLpq{\modeldist(\vx|\vz)p(\vz)}{p_{\gamma}(\vz|\vx)\tdistx}$, VEEGAN uses the density ratio trick - and hence also loses an estimate for the bound -- Appendix~\ref{app:veegan_bound_tracking}. Since the expectation in the objective is not taken with respect to the data distribution $\tdistx$ but the joint distribution $\modeldist(\vx|\vz) p(\vz)$ decoder gradients follow from the variational objective.
\begin{figure*}[t!]
	\centering
	\captionsetup[subfigure]{labelformat=empty}
	\hspace{-7mm}
	\begin{subfigure}{0.12\textwidth}{
			\centering
			\includegraphics[height=2.8cm]{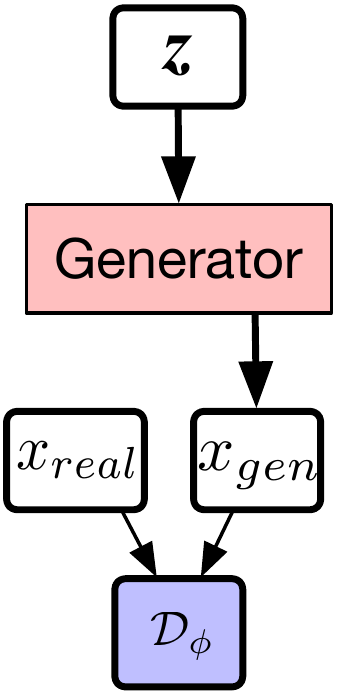}
			\caption{DCGAN}
	}\end{subfigure}
	\hspace{1mm}
	\begin{subfigure}{0.2\textwidth}{
			\centering
			\includegraphics[height=2.8cm]{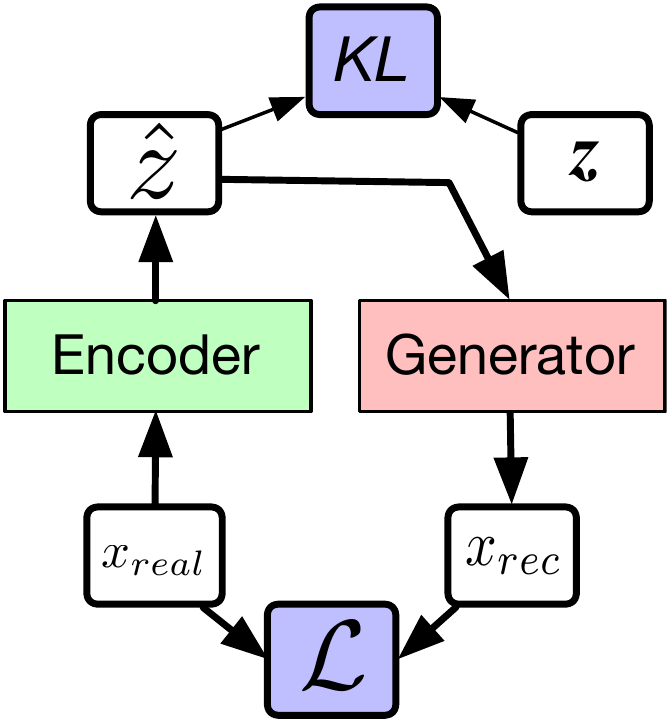}
			\caption{VAE}
	}\end{subfigure}
	\hspace{1mm}
	\begin{subfigure}{0.19\textwidth}{
			\centering
			\includegraphics[height=2.8cm]{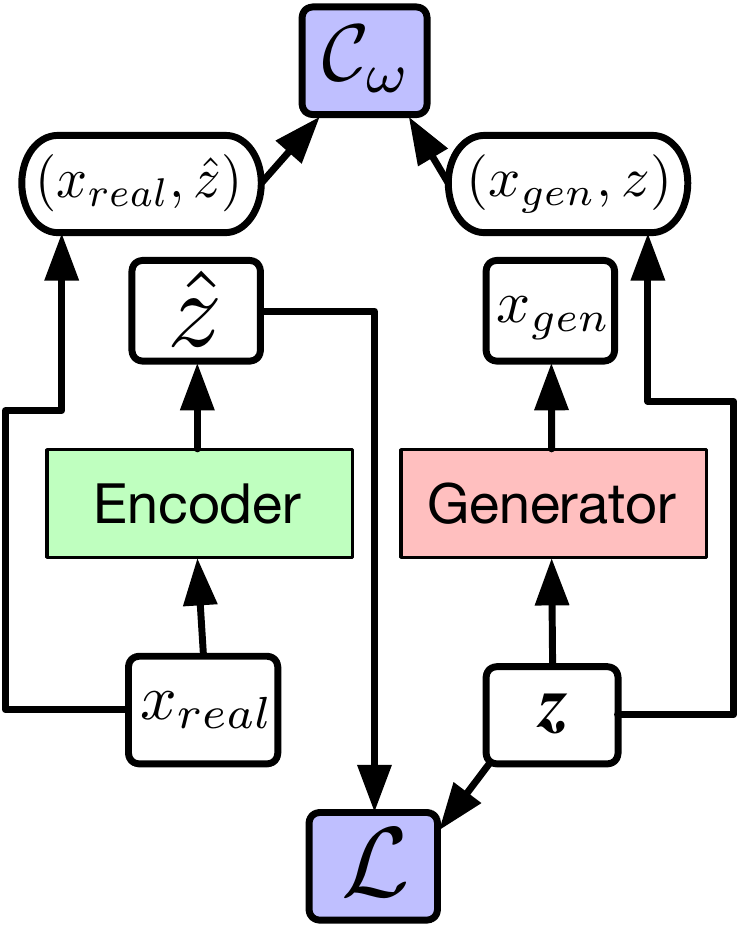}
			\caption{VEEGAN}
	}\end{subfigure}
	\hspace{1mm}
	\begin{subfigure}{0.2\textwidth}{
			\centering
			\includegraphics[height=2.8cm]{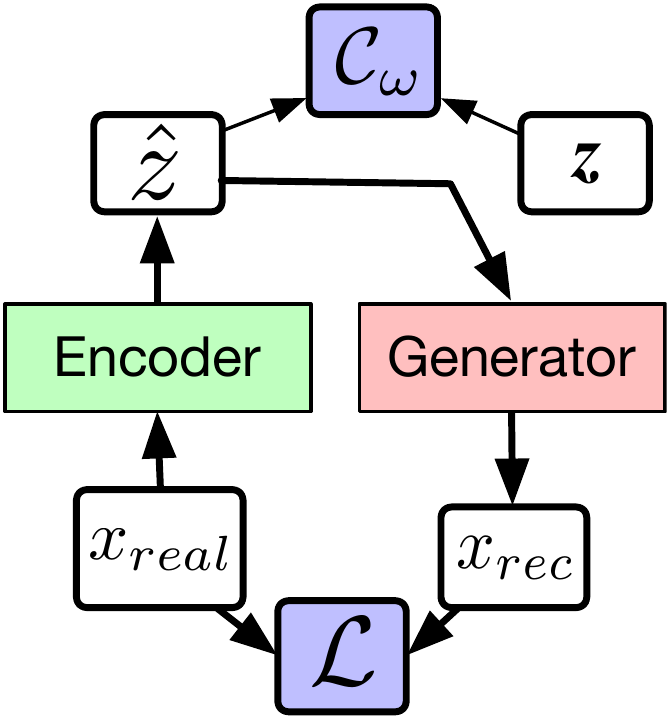}
			\caption{AAE}
	}\end{subfigure}
	\hspace{1mm}
	\begin{subfigure}{0.2\textwidth}{
			\label{fig:model_our}
			\centering
			\includegraphics[height=2.8cm]{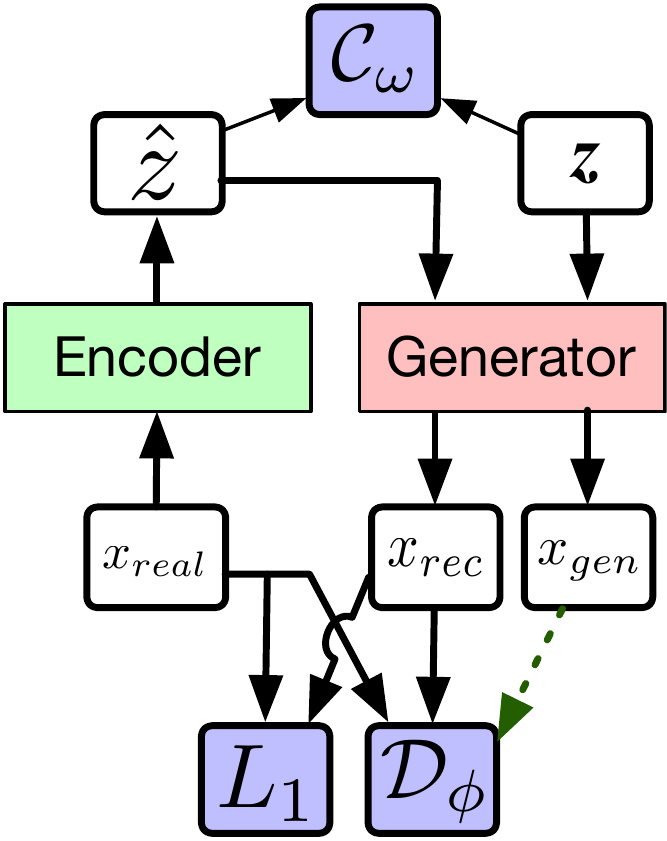}
			\caption{\reconourgan/\ourgan}
	}\end{subfigure}
	\caption{{Model architectures (WGAN is similar to DCGAN). The difference
			between~\ourgan~and~\reconourgan~is exemplified using the {\color{Green}green arrow} between
			$x_{gen}$ and $\disc$ .}}
	\label{fig:model}
\end{figure*}

\section{EVALUATION}
We compare marginal distribution matching using implicit distributions in variational inference provided by
VAE-GAN hybrids with VAEs and GANs on image generation tasks.
We perform an extensive comparison between (detailed in Figure~\ref{fig:model}):
\vspace{-0.8em}
\begin{itemize}[leftmargin=*,noitemsep]
\item VAEs with diagonal Gaussian posteriors -- the most widespread variational inference model.
\item DCGAN and WGAN-GP to provide a baseline for VAE-GAN hybrids.
\item AAE to exhibit the effect of replacing the analytical KL in VAEs with the density ratio trick.
\item VEEGAN to evaluate the density ratio trick on joint distributions, and the effect of reconstruction losses in latent space rather than data space.
\item \reconourgan~and~\ourgan, two Variational GAN Hybrids we introduce inspired by the analysis in Section~\ref{sec:synth_like}. Like AAEs, these models use marginal distribution matching in latent space via the density ratio trick. In visible space, \reconourgan~uses a discriminator trained on reconstructions (Equation~\ref{eq:gan_d_recon_ours}), and ~\ourgan~ uses a discriminator trained on both data and samples (Equation~\ref{eq:gan_d_ours}). Both use an $l_1$ reconstruction loss in data space. Details are provided in Appendix~\ref{alg:pseudocode}. Contrasting ~\reconourgan~ and AAEs measures the effect of explicit distribution matching in visible space, while comparing~\reconourgan~and~\ourgan~assesses the efficacy of the density ratio trick in latent space.
\end{itemize}

We do not compare with AdversarialVB, as it underperformed in our initial experiments
(Appendix~\ref{app:adv_vb}). Other approaches to VAE-GAN hybrids include replacing reconstruction
losses on pixels with reconstruction losses on discriminator features \citep{aepixels}, but they are outside
the scope of this work.

We train models on \colormnist, CelebA and CIFAR-10 and complement visual inspection of samples with three metrics: Inception Score, sample diversity, independent Wasserstein critic. We do not use the ELBO as an evaluation metric, since we have shown VAE-GAN hybrids cannot estimate it reliably. Descriptions of the metrics, experimental details and samples are in Appendices~\ref{sec:metrics},~\ref{app:hyperparameters} and ~\ref{app:samples}.

\begin{figure*}[ht!]
\begin{center}
\begin{subfigure}{0.32\textwidth}{
\includegraphics[width=\textwidth]{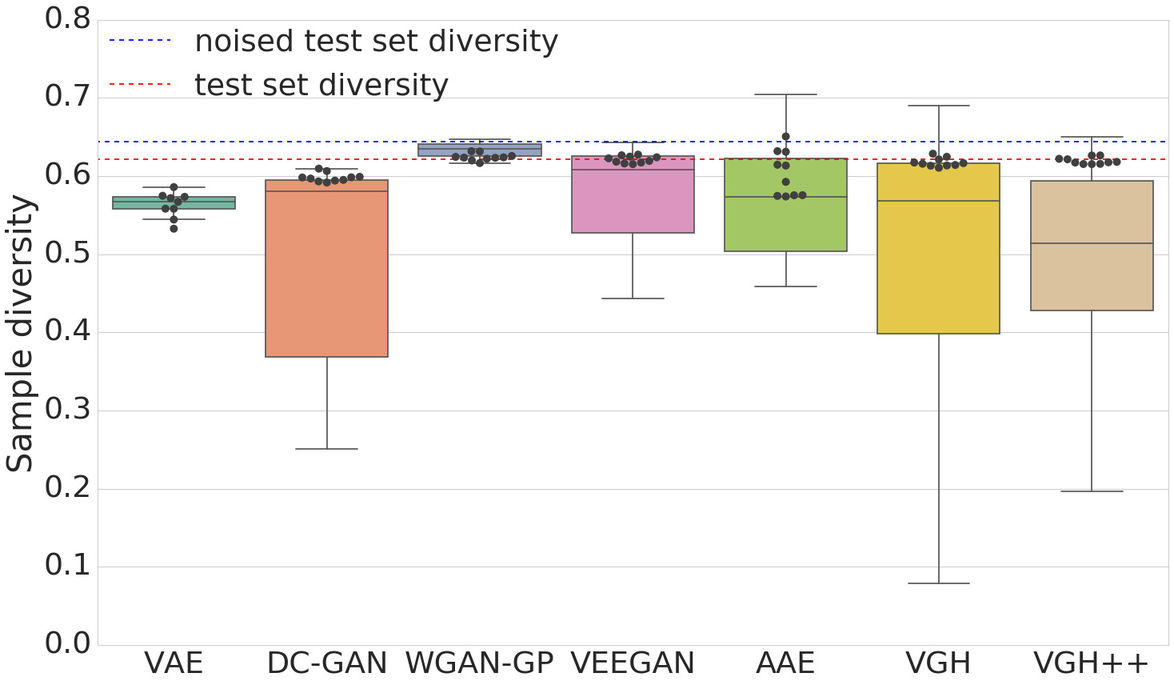}
\caption{Diversity score (CelebA)}
\label{fig:mssim_inception:celeba}
}\end{subfigure}
\begin{subfigure}{0.32\textwidth}{
\includegraphics[width=\textwidth]{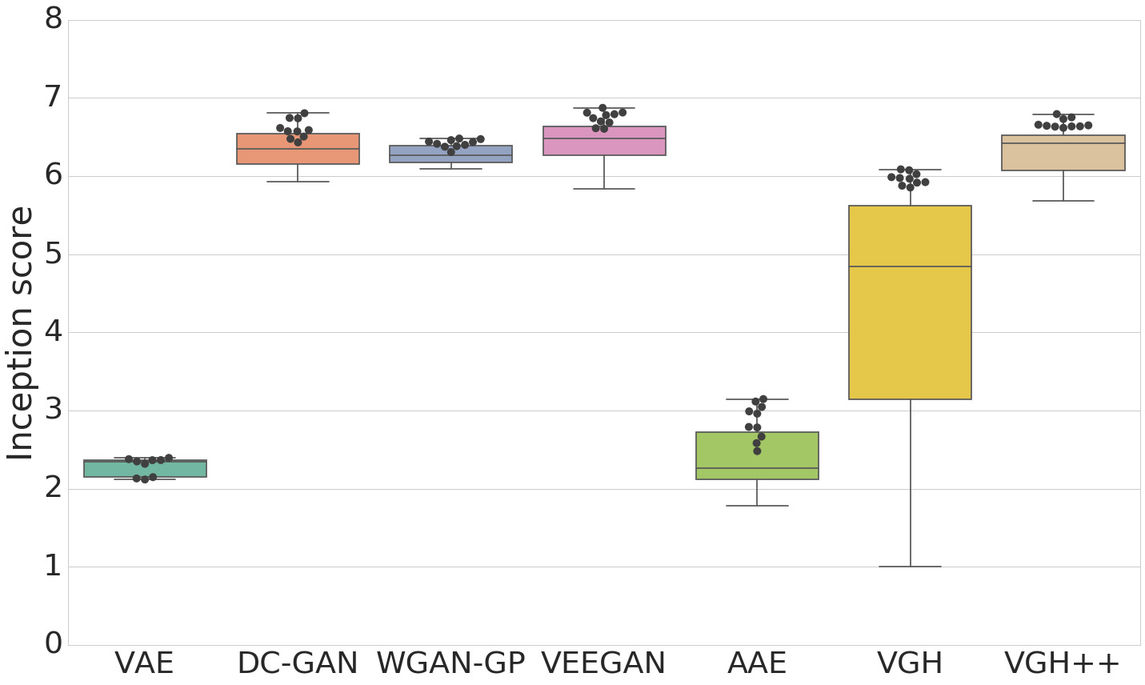}
\caption{Inception score (ImageNet)}
\label{fig:mssim_inception:inceptionnet}
}\end{subfigure}
\begin{subfigure}{0.32\textwidth}{
\includegraphics[width=\textwidth]{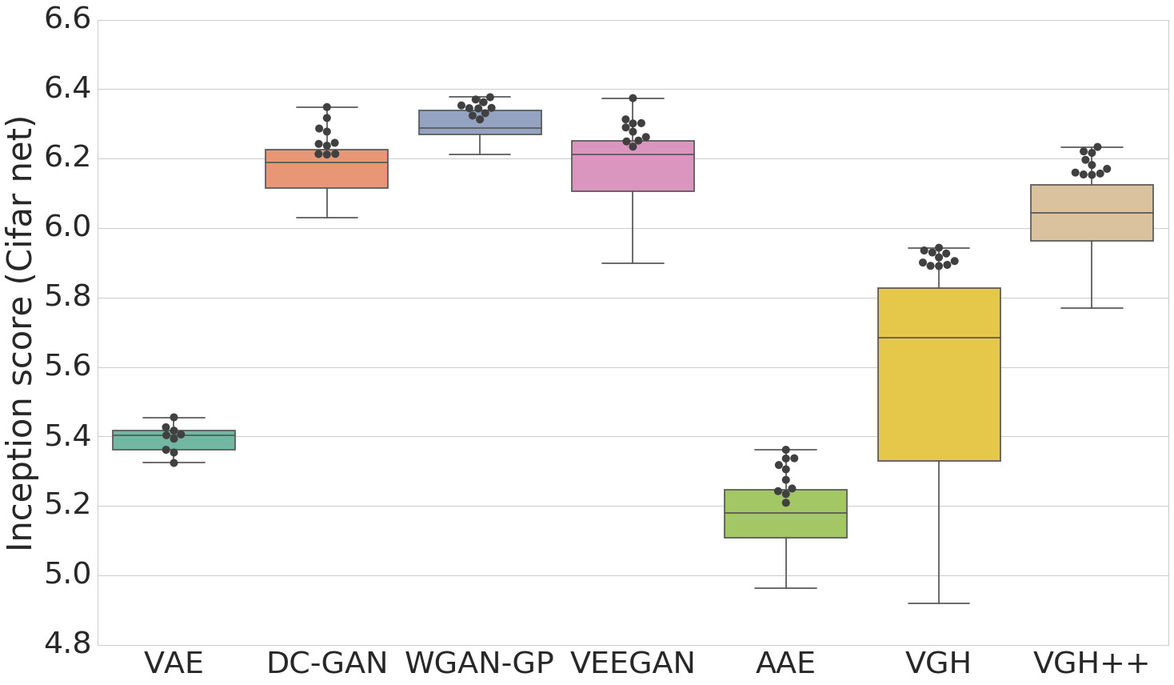}
\caption{Inception score (CIFAR)}
\label{fig:mssim_inception:cifarnet}
}\end{subfigure}
\caption{
(Left) Sample diversity on CelebA, and is viewed relative to test set: too much
diversity shows failure to capture the data distribution, too little is indicative of mode collapse. We also
report the diversity obtained on a noised-version of the test set, which has a higher diversity than the
test set. (Middle) Inception scores on CIFAR-10. (Right) Inception scores computed using a VGG-style
network on CIFAR-10. For inception scores, higher values are better. For test data, diversity score: 0.621,
inception score: 11.25, inception score (using CIFAR-10 trained net): 9.18. Best results are shown with
black dots, and box plots show the hyperparamter sensitivity.}
\label{fig:mssim_inception}
\end{center}
\vspace{-2mm}
\end{figure*}

\begin{figure*}[ht!]
\begin{center}
\begin{subfigure}{0.32\textwidth}{
\includegraphics[width=\textwidth]{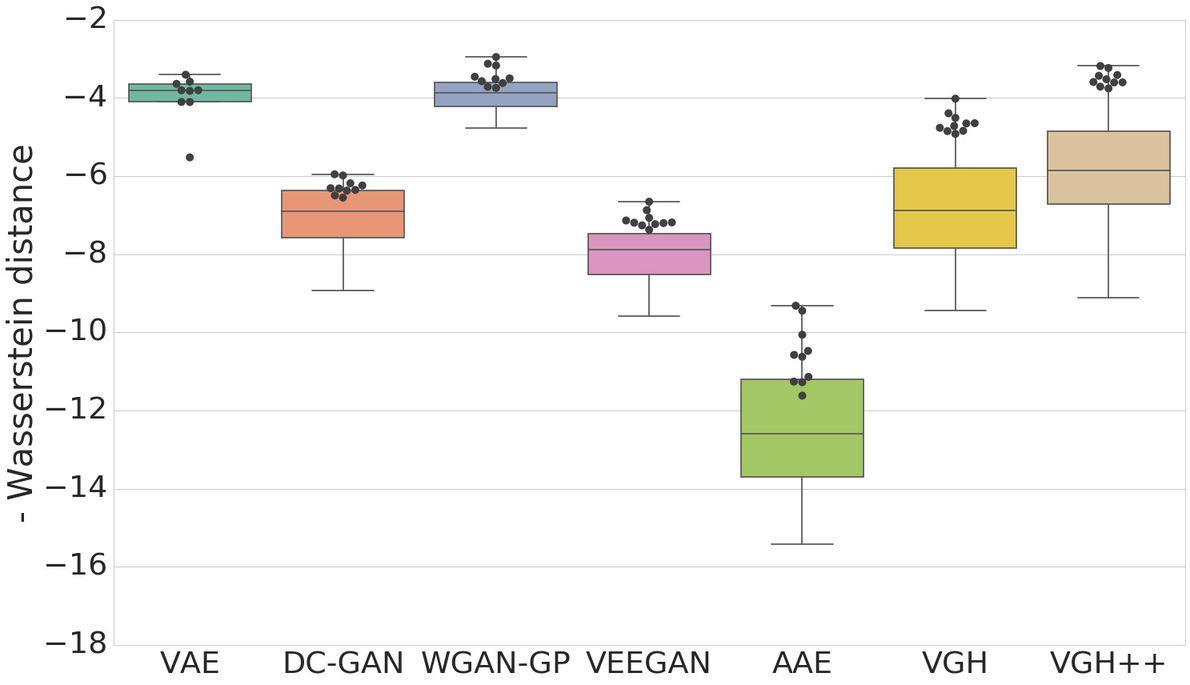}
\caption{\colormnist}
\label{fig:wcritics-colormnist}
}\end{subfigure}
\begin{subfigure}{0.32\textwidth}{
\includegraphics[width=\textwidth]{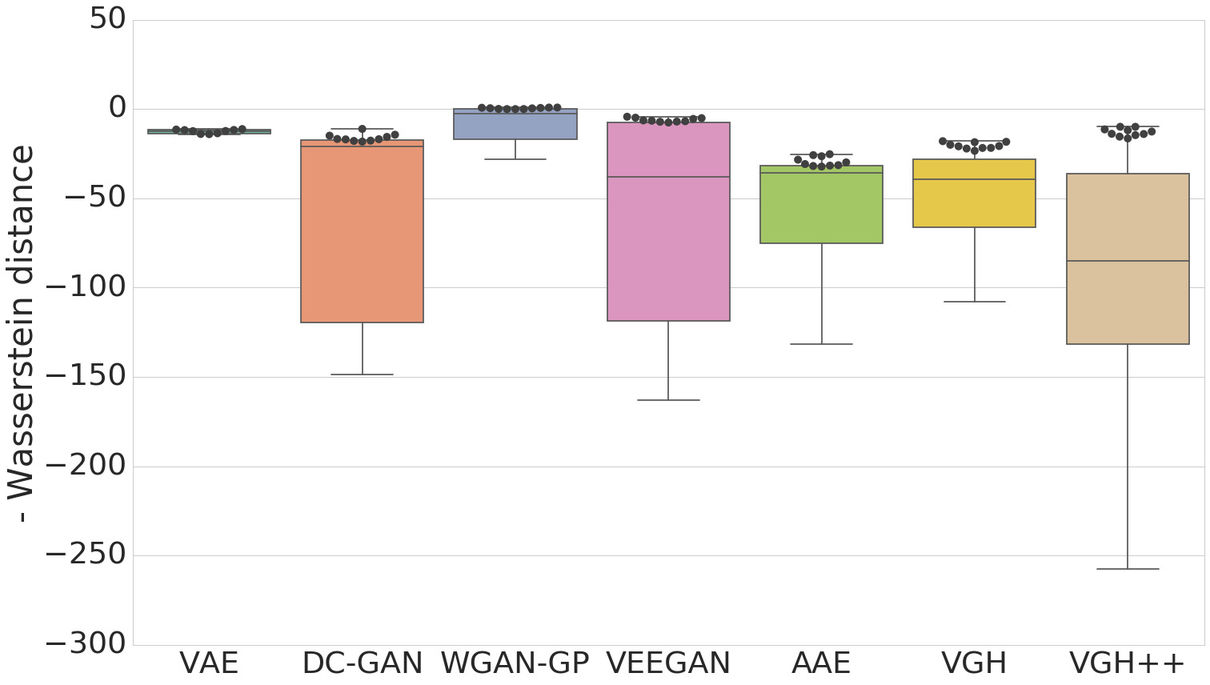}
\caption{CelebA}
\label{fig:wcritics-celeba}
}\end{subfigure}
\begin{subfigure}{0.32\textwidth}{
\includegraphics[width=\textwidth]{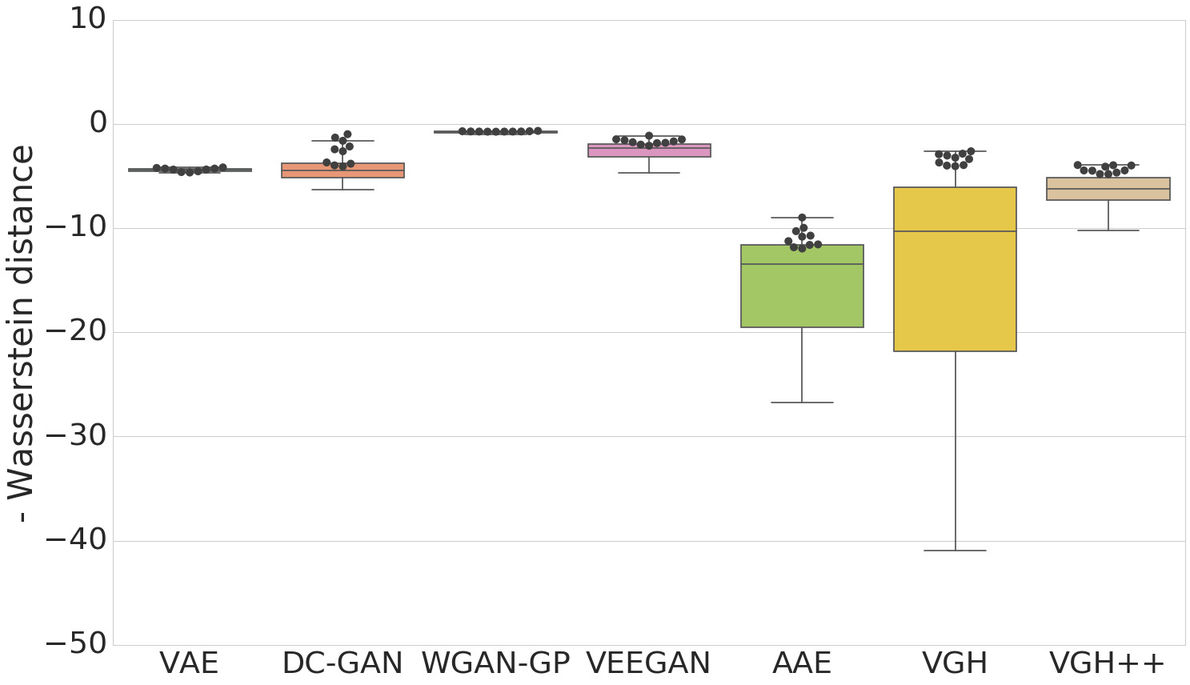}
\caption{CIFAR-10}
\label{fig:wcritics-cifar}
}\end{subfigure}
\caption{Comparison using negative Wasserstein distance computed using an independent Wasserstein
critic. Higher is better. The metric captures overfitting and low quality samples and been shown to
correlate
with human evaluations \citep{jiwoong2018quantitatively}.}
\label{fig:wcritics-all}
\end{center}
\end{figure*}

Figures~\ref{fig:mssim_inception}~and~\ref{fig:wcritics-all} show that VAEs perform well on datasets that have less variability, such as ColorMNIST and CelebA, but are not able to capture the subtleties of CIFAR10. Marginal distribution matching in visible space improves generation quality, with VEEGAN,~\ourgan~and~\reconourgan~performing better than VAEs and AAEs. However, VAE-GAN hybrids do not outperform GANs on image quality metrics and consistently exhibit a higher sensitivity to hyperparameters, caused by additional optimization components. \ourgan~ consistently outperforms~\reconourgan, showing that the density ratio trick has issues matching the marginal latent posterior to the prior and that matching marginal distributions in visible space explains the increased sample quality of VAE-GAN hybrids compared to VAEs.

By assessing sample quality and diversity, our experiments show that, at present, merging variational inference with implicit and marginal distribution matching does not provide a clear benefit.

\section{CONCLUSION}
\vspace{-4.2mm}
We have shown the widespread failure to learn marginal distributions with Variational Autoencoders. We
asked whether this is the effect of conditional distribution matching, and of explicit posteriors and explicit
model distributions. To test this hypothesis, we explored marginal distribution matching and implicit
distribution in variational inference, through existing and new VAE-GAN hybrids.

Through a wide range of experiments, we have shown that VAE-GAN hybrids do not deliver on the promise of addressing major challenges in variational inference. Problems with value estimation of divergences caused by the use of classifier probabilities, difficulties of scaling to high dimensions, hyperparameter sensitivity and the struggles to outperform GANs on sample quality metrics, limits, at present, the applicability of VAE-GAN hybrids. Since implicit models and adversarial training do not solve the obstacles of variational inference, distribution matching in latent and visible space remain important generative models research issues.

\clearpage
\newpage

\vspace{-5mm}
\balance
\bibliography{paper}%
\bibliographystyle{abbrvnat}

\clearpage
\appendix
\begin{large}
\textbf{Appendix --\ourtitle}
\end{large}

\section{Low posterior VAE samples}\label{app:breaking_vaes}

\begin{figure*}[htbp]
\begin{center}
\begin{subfigure}{0.95\textwidth}{
\centering
\includegraphics[width=0.41\textwidth]{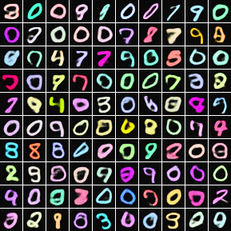} \addhspace
\includegraphics[width=0.41\textwidth]{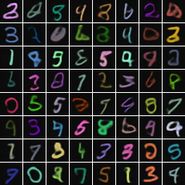} \addhspace
\caption{ColorMNIST}
\label{fig:cmnist:bad_samples_alongside}
}\end{subfigure}
\begin{subfigure}{0.95\textwidth}{
\centering
\includegraphics[width=0.41\textwidth]{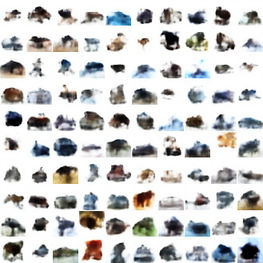} \addhspace
\includegraphics[width=0.41\textwidth]{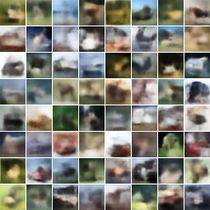} \addhspace
\label{fig:cifar:bad_samples_alongside}
\caption{CIFAR-10}
}\end{subfigure}
\begin{subfigure}{0.95\textwidth}{
\centering
\includegraphics[width=0.41\textwidth]{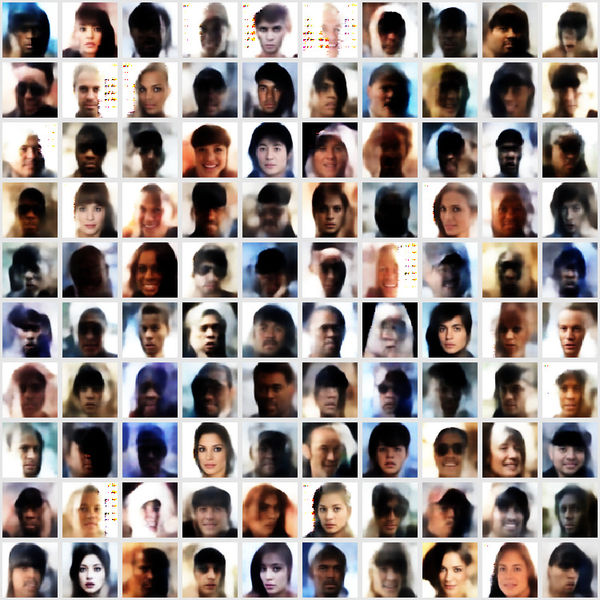} \addhspace
\includegraphics[width=0.41\textwidth]{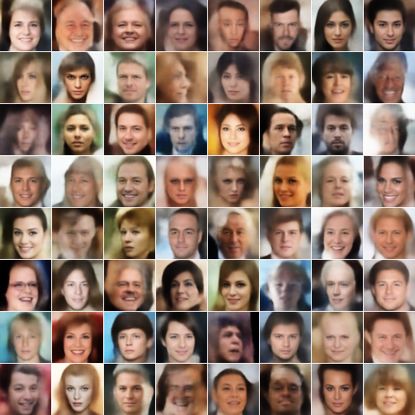} \addhspace
\caption{CelebA}
\label{fig:celeba:bad_samples_alongside}
}\end{subfigure}
\caption{
Low posterior VAE samples obtained by sampling latent variables from an area in the latent space where the marginal posterior has low probability mass (left), alongside standard VAE samples from the \textbf{same} model (right).
}
\label{fig:bad_samples_alongside}
\end{center}
\end{figure*}

In this section we detail the experiments performed to create low posterior VAE samples. For the algorithm used to obtained the low posterior samples, see Algorithm~\ref{alg:pseudocode_generate_low_post_samples}. For the low posterior samples alongside standard samples from the model, see Figure~\ref{fig:bad_samples_alongside}.

\begin{algorithm}
\caption{Pseudocode for generating low posterior VAE samples}
\label{alg:pseudocode_generate_low_post_samples}
\begin{algorithmic}[1]
\State Load trained variational model with prior $\zdist$, and posterior $\vardistz$.
\State $\mathsf{log\_q\_to\_z} = \{\}$
\For{$\mathsf{i}=1:\mathsf{num\_z}$}
  \State sample $\vz_{i}$ from $\zdist$
  \State $\mathsf{posterior\_list} = []$
  \For{$\mathsf{\vx}$ in dataset}
      \State append $\log \vardistz (\vz_{i})$ to $\mathsf{posterior\_list}$
  \EndFor
  \State $\log q(\vz_{i}) = \mathsf{log\_mean\_exp} (\mathsf{posterior\_list})$
  \State $\mathsf{log\_q\_to\_z}[\log q(\vz_{i})] = \vz_i$
\EndFor
\State Sort $\mathsf{log\_q\_to\_z}$ by key and let $\mathsf{z\_adv}$ be the list of values corresponding to the $n$ smallest keys.

\State $\mathsf{low\_posterior\_samples} = []$
\For{$\vz$ in $\mathsf{z\_adv}$}
  \State append a sample from $\modeldist(\vx|\vz)$ to $\mathsf{low\_posterior\_samples}$
\EndFor
\State return $\mathsf{low\_posterior\_samples}$
\end{algorithmic}
\end{algorithm}

\subsection{Low posterior samples - model analysis}
\begin{figure*}[htbp]
\begin{center}
\begin{subfigure}{\textwidth}{
\includegraphics[width=0.3\textwidth]{cmnist_bad_samples} \addhspacesmall
\includegraphics[width=0.3\textwidth]{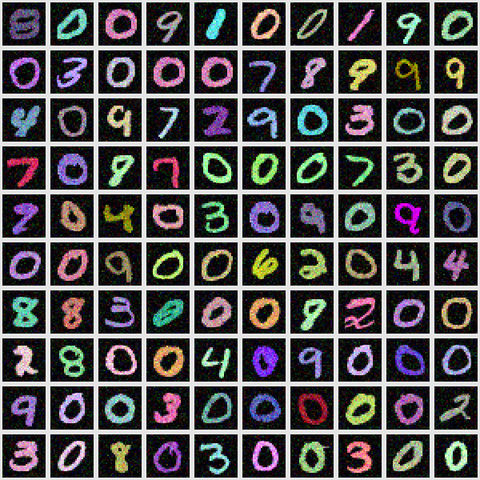} \addhspacesmall
\includegraphics[width=0.3\textwidth]{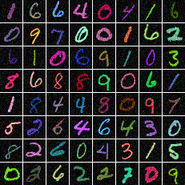} \addhspacesmall
\caption{ColorMNIST}
\label{fig:cmnist:bad_samples_nn}
}\end{subfigure}
\begin{subfigure}{\textwidth}{
\includegraphics[width=0.3\textwidth]{cifar_bad_samples} \addhspacesmall
\includegraphics[width=0.3\textwidth]{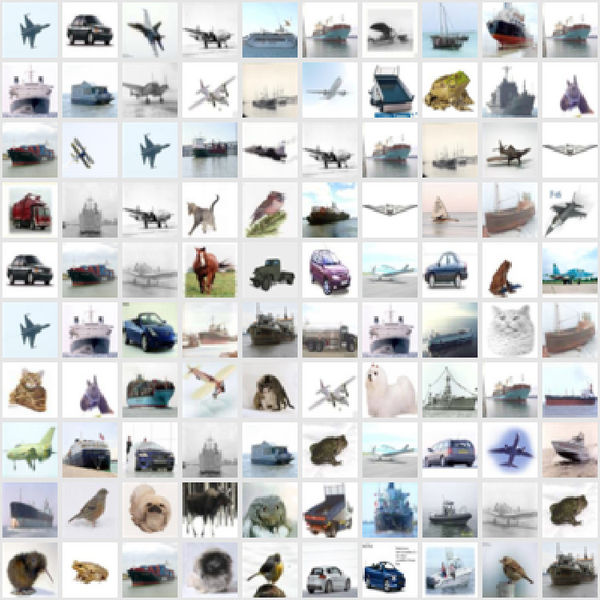} \addhspacesmall
\includegraphics[width=0.3\textwidth]{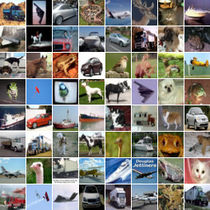} \addhspacesmall
\caption{CIFAR-10}
\label{fig:cifar:bad_samples_nn}
}\end{subfigure}
\begin{subfigure}{\textwidth}{
\includegraphics[width=0.3\textwidth]{celeba_bad_samples} \addhspacesmall
\includegraphics[width=0.3\textwidth]{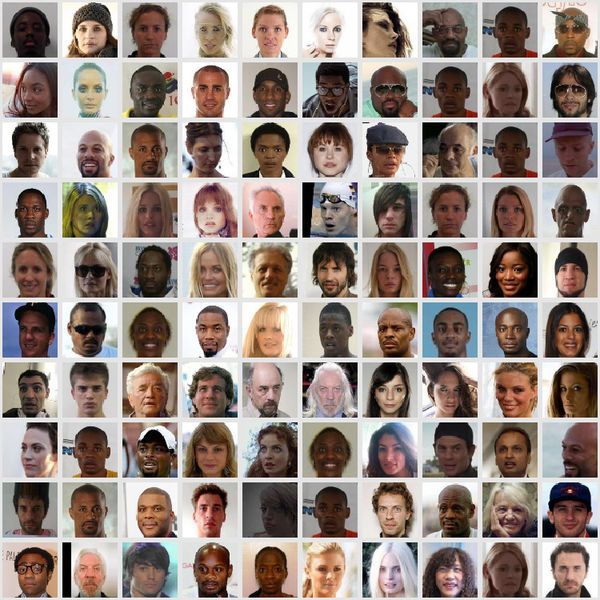} \addhspacesmall
\includegraphics[width=0.3\textwidth]{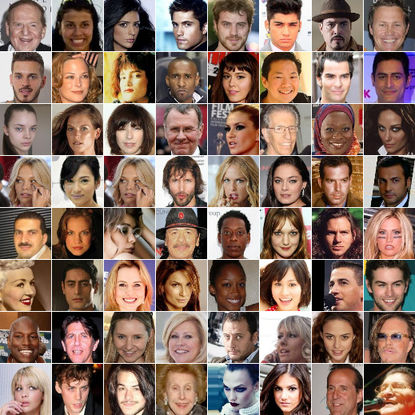} \addhspacesmall
\caption{CelebA}
\label{fig:celeba:bad_samples_nn}
}\end{subfigure}
\caption{Low posterior samples (left), the nearest neighbors in the dataset from the low posterior samples (middle), uniformly dataset examples (right).}
\label{fig:nn_bad_samples}
\end{center}
\end{figure*}

\begin{figure*}[htbp]
\begin{center}
\begin{subfigure}{0.3\textwidth}{
\includegraphics[width=\textwidth]{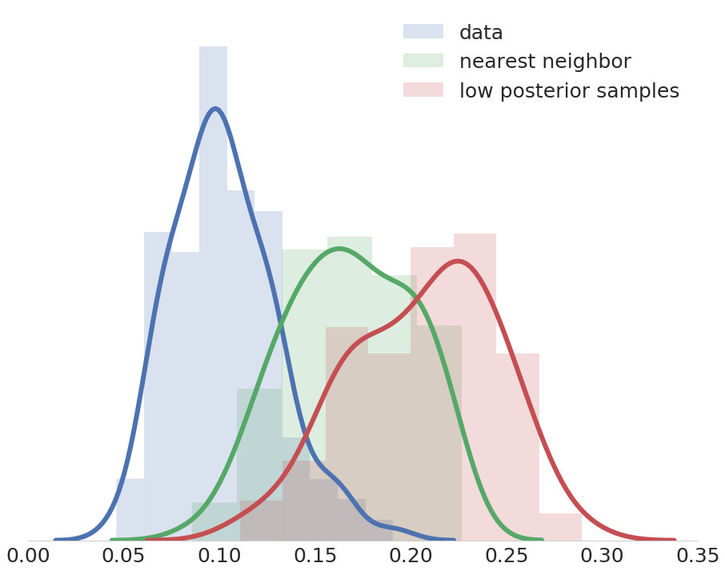} \addhspacesmall
\caption{ColorMNIST}
\label{fig:cmnist:bad_samples_pixels}
}\end{subfigure}
\begin{subfigure}{0.3\textwidth}{
\includegraphics[width=\textwidth]{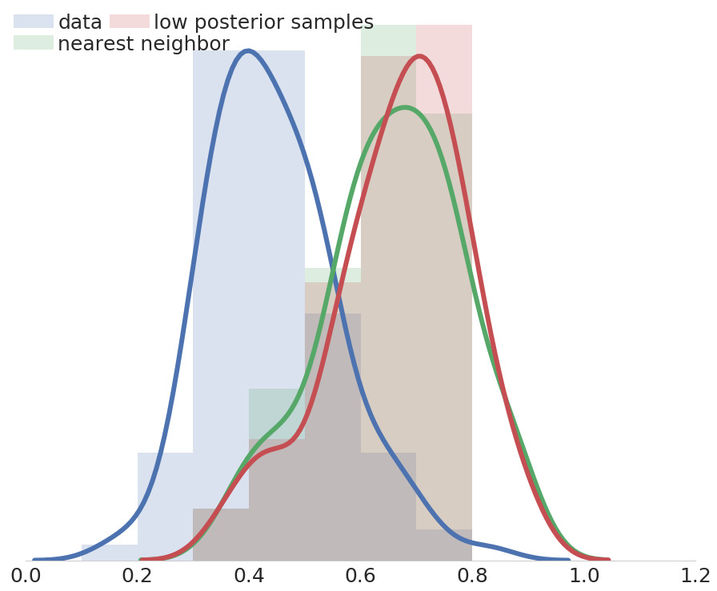} \addhspacesmall
\caption{CIFAR-10}
\label{fig:cifar:bad_samples_pixels}
}\end{subfigure}
\begin{subfigure}{0.3\textwidth}{
\includegraphics[width=\textwidth]{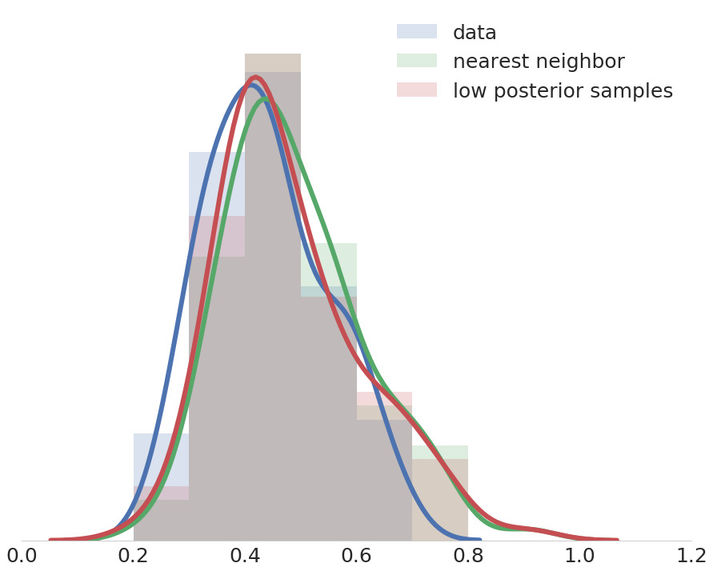} \addhspacesmall
\caption{CelebA}
\label{fig:celeba:bad_samples_pixels}
}\end{subfigure}
\caption{Histogram of pixels on uniformly sampled data, the nearest neighbors from the low posterior samples and the low posterior samples. For CIFAR-10 and ColorMNIST we see that both the low posterior samples and their nearest neighbors in the dataset are atypical compared to the data. We also plot the result of using a KDE density estimation for each histogram.}
\label{fig:pixels_bad_samples}
\end{center}
\end{figure*}

\begin{figure*}[htbp]
\begin{center}
\begin{subfigure}{0.3\textwidth}{
\includegraphics[width=\textwidth]{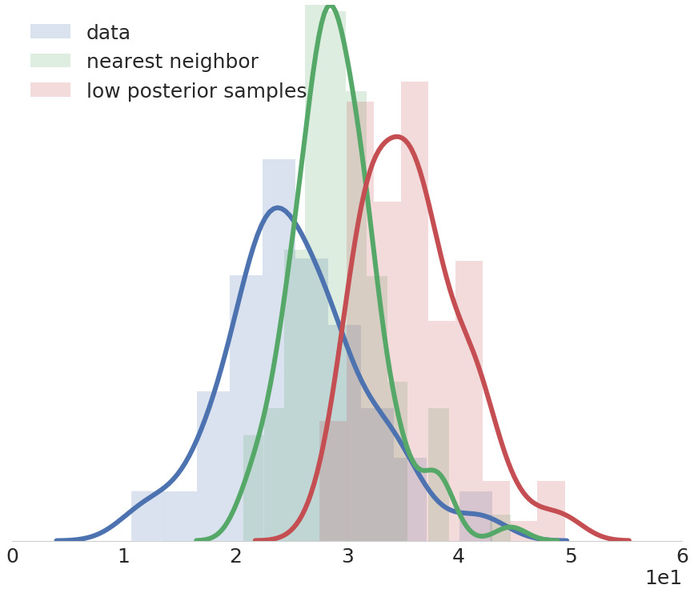} \addhspacesmall
\caption{ColorMNIST}
\label{fig:cmnist:bad_samples_kl}
}\end{subfigure}
\begin{subfigure}{0.3\textwidth}{
\includegraphics[width=\textwidth]{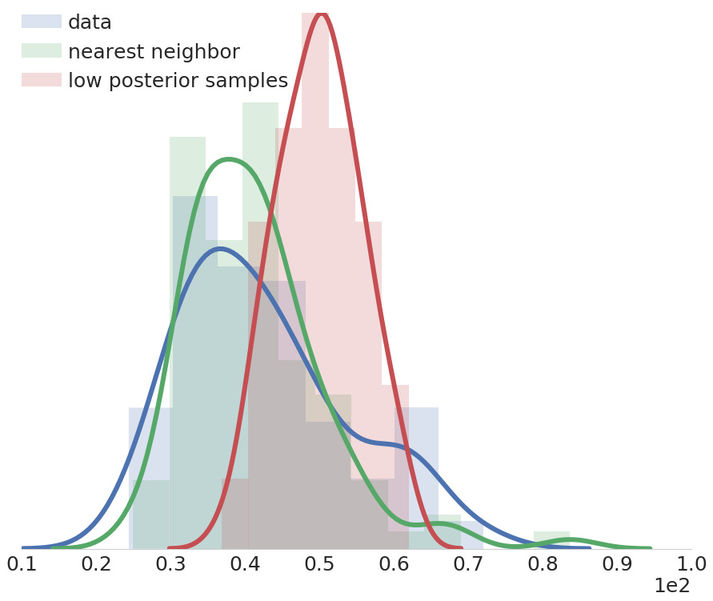} \addhspacesmall
\caption{CIFAR-10}
\label{fig:cifar:bad_samples_kl}
}\end{subfigure}
\begin{subfigure}{0.3\textwidth}{
\includegraphics[width=\textwidth]{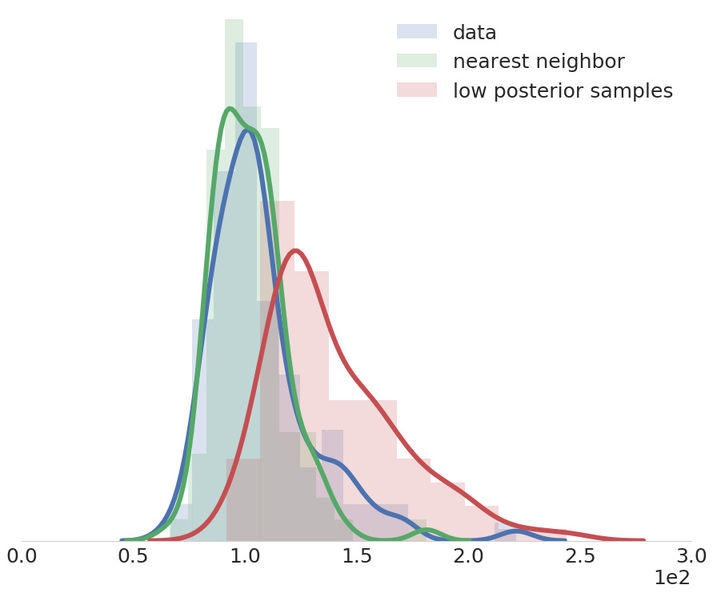} \addhspacesmall
\caption{CelebA}
\label{fig:celeba:bad_samples_kl}
}\end{subfigure}
\caption{Histogram of $\KLpq{\vardistzn}{\zdist}$ on uniformly sampled data, the nearest neighbors from the low posterior samples and the low posterior samples. Overall, we see a higher KL term for data points close to the low posterior samples and for low posterior samples. We also plot the result of using a KDE density estimation for each histogram.}
\label{fig:kl_bad_samples}
\end{center}
\end{figure*}

\begin{figure*}[htbp]
\begin{center}
\begin{subfigure}{0.3\textwidth}{
\includegraphics[width=\textwidth]{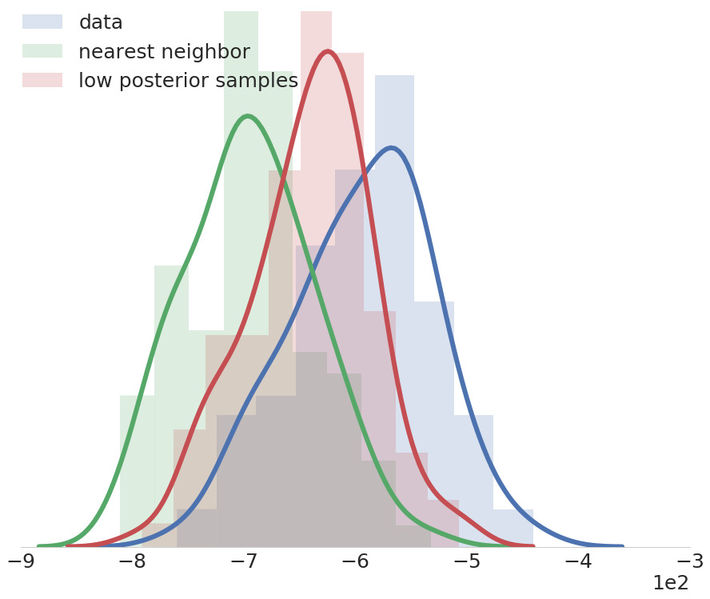} \addhspacesmall
\label{fig:cmnist:bad_samples_elbo_app}
\caption{ColorMNIST}
}\end{subfigure}
\begin{subfigure}{0.3\textwidth}{
\includegraphics[width=\textwidth]{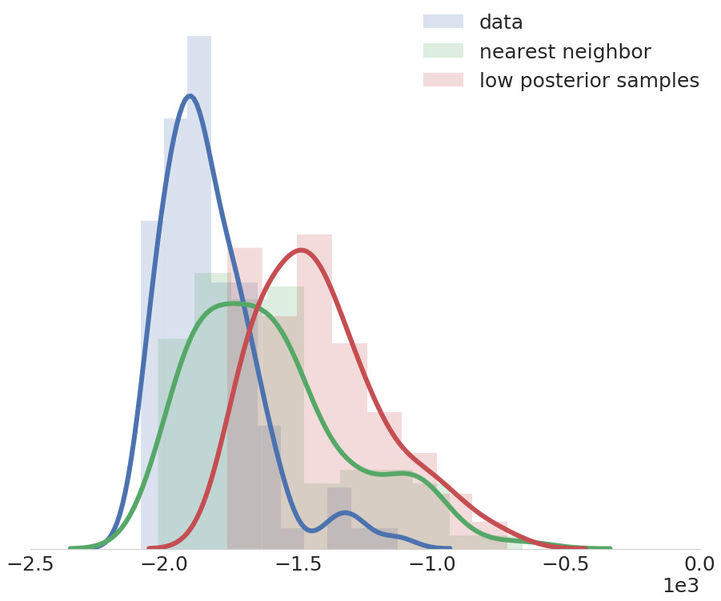} \addhspacesmall
\caption{CIFAR-10}
\label{fig:cifar:bad_samples_elbo_app}
}\end{subfigure}
\begin{subfigure}{0.3\textwidth}{
\includegraphics[width=\textwidth]{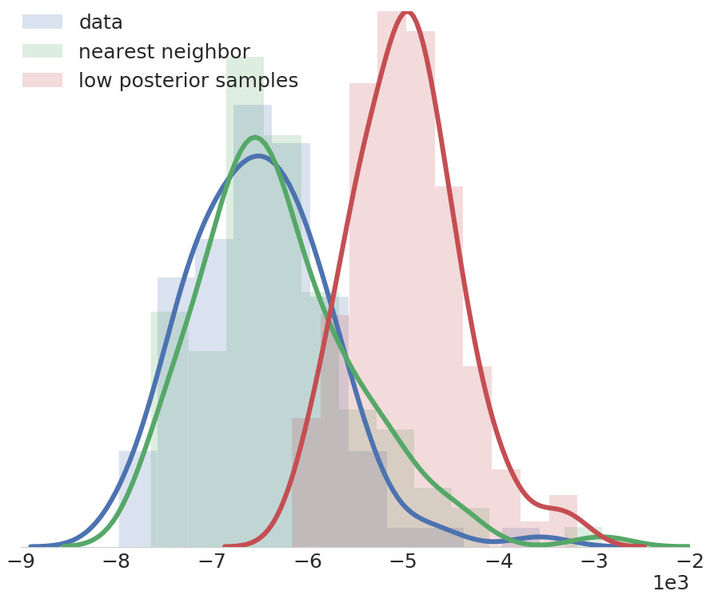} \addhspacesmall
\caption{CelebA}
\label{fig:celeba:bad_samples_elbo_app}
}\end{subfigure}
\caption{Model evidence lower bound of uniformly sampled data, the nearest neighbors from the low posterior samples in the dataset and the low posterior samples. While on ColorMNIST the model recognizes that the low posterior samples and their nearest neighbors have a lower likelihood than the data, for CIFAR-10 and CelebA the model thinks the low posterior samples are more likely than the data. We also plot the result of using a KDE density estimation for each histogram.}
\label{fig:elbo_bad_samples_app}
\end{center}
\end{figure*}

In order to understand why the low posterior samples look as shown in Figure~\ref{fig:bad_samples_alongside}, we performed an analysis to show how these samples compare to the data distribution. For ColorMNIST, we visually saw that the samples are thicker than the data or the standard samples, while for CelebA and CIFAR-10 we saw predominantly white backgrounds so we plotted the pixel histogram of the dataset against the pixel histogram of the low posterior samples (Figure~\ref{fig:pixels_bad_samples}). The histograms of pixels for data, low posterior samples and their nearest neighbors in the dataset shows that the low posterior samples differ in pixel composition compared to the uniformly sampled data, but to check whether images like this exist in the dataset we plot the nearest neighbors in $l_2$ distance from the dataset to the low posterior samples (see Figure~\ref{fig:nn_bad_samples}). This analysis shows that data examples that are similar to the low posterior samples exist, but based on the histogram analysis and visual inspection we know that they have low probability under the true data distribution. A hypothesis emerges: the model is not putting mass in the marginal posterior distribution for areas of the space that encode data points which have low probability under the true data distribution. To test this hypothesis, we report the histograms of average $\KLpq{\vardistzn}{\zdist}$, obtained by encoding data sampled from the dataset uniformly compared to the nearest neighbors of the low posterior samples Engineering Retreat Googl examples and the low posterior samples themselves (Figure~\ref{fig:kl_bad_samples}). Our results show that indeed, the data points closest to the low posterior samples have a higher KL cost compared to datapoints sampled uniformly from the dataset and that these data points are unlikely under the data distribution. Knowing that these examples are unlikely under the true data distribution, we expect to see the same under the \textit{model} distribution. In Figure~\ref{fig:elbo_bad_samples_app} we show that for CIFAR-10 and CelebA, the model reports that the low posterior samples are more likely than the data. This demonstrated that the model is unable to capture the subtleties of the data distribution, and can be fooled into predicting high likelihoods for samples that have low probability under the true data distribution. In this work we have exploited the gap between the prior and marginal posterior distributions in VAEs trained with Gaussian posteriors, to show that the model can generate samples are far from the sample distribution and far from the data distribution. Previous work has focused on finding adversarial examples as input to the VAE by finding points in data space that the VAE is unable to reconstruct \citep{kos2017adversarial}.

\begin{figure*}[htbp]
\begin{center}
\begin{subfigure}{0.32\textwidth}{
\includegraphics[width=0.95\textwidth]{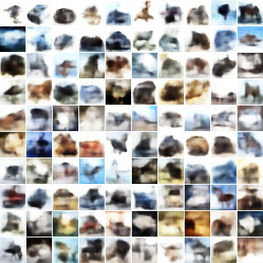}
\label{fig:cifar:bad_samples_5k_z}
\caption{5000}
}\end{subfigure}
\begin{subfigure}{0.32\textwidth}{
\includegraphics[width=0.95\textwidth]{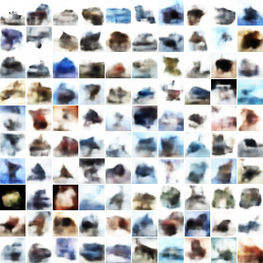}
\label{fig:cifar:bad_samples_50k_z}
\caption{50000}
}\end{subfigure}
\begin{subfigure}{0.32\textwidth}{
\includegraphics[width=0.95\textwidth]{cifar_bad_samples}
\label{fig:cifar:bad_samples_many_z}
\caption{5000000}
}\end{subfigure}
\caption{Low posterior samples on CIFAR-10 resulting from different numbers of latents sampled from the prior (corresponding to $\mathsf{num\_z}$ in Algorithm~\ref{alg:pseudocode_generate_low_post_samples}). While the samples get more pathological with an increased number of samples from the prior, we can already generate abnormal VAE samples from a small number of latent samples.}
\label{fig:pixels_different_count}
\end{center}
\end{figure*}

\section{Scaling up latent spaces in VAEs and AAEs}
\label{app:scaling}
\begin{figure*}[htbp]
\begin{center}
\begin{subfigure}{0.32\textwidth}{
\includegraphics[width=0.95\textwidth]{data_cmnist}
\caption{Data}
}\end{subfigure}
\begin{subfigure}{0.32\textwidth}{
\includegraphics[width=0.95\textwidth]{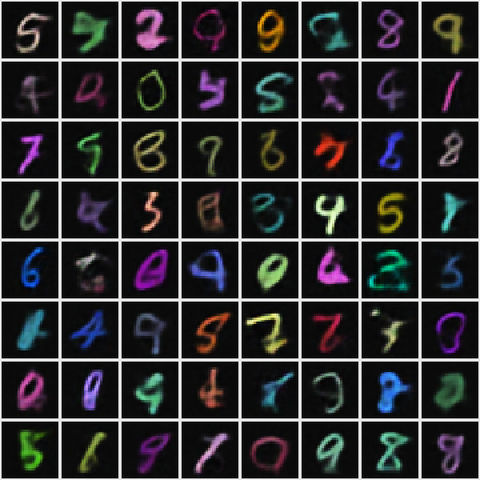}
\label{fig:va_samples_10k_latents}
\caption{VAE samples}
}\end{subfigure}
\begin{subfigure}{0.32\textwidth}{
\includegraphics[width=0.95\textwidth]{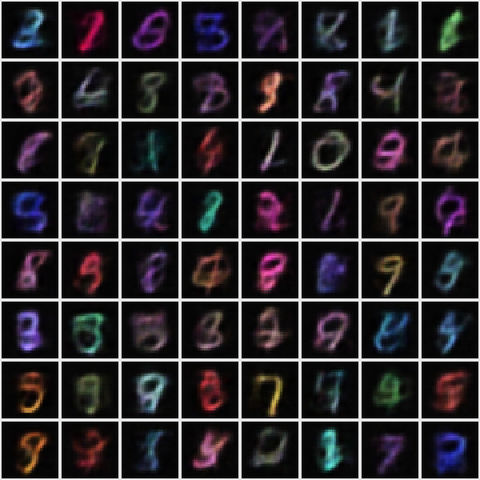}
\label{fig:aae_samples_10k_latents}
\caption{AAE samples}
}\end{subfigure}
\caption{Samples from VAE and AAE with 10k latents compared to data on \colormnist.}
\label{fig:samples_10k}
\end{center}
\end{figure*}

Figure~\ref{fig:samples_10k} shows that VAEs scale better than AAEs to a high number of latent size - in this case 10000.

\begin{figure*}[htbp]
\begin{center}
\begin{subfigure}{0.32\textwidth}{
\includegraphics[width=0.95\textwidth]{data_celeba}
\caption{Data}
}\end{subfigure}
\begin{subfigure}{0.32\textwidth}{
\includegraphics[width=0.95\textwidth]{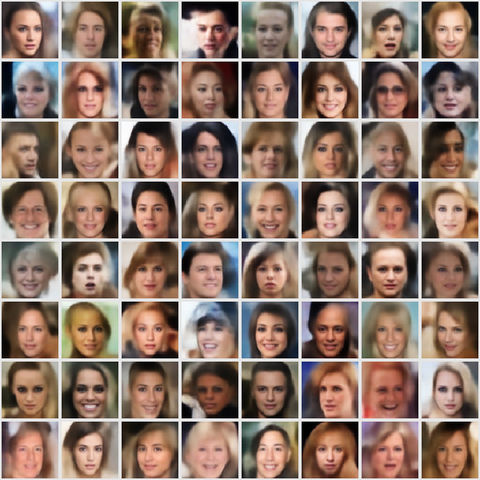}
\label{fig:va_samples_8k_latents}
\caption{VAE samples}
}\end{subfigure}
\begin{subfigure}{0.32\textwidth}{
\includegraphics[width=0.95\textwidth]{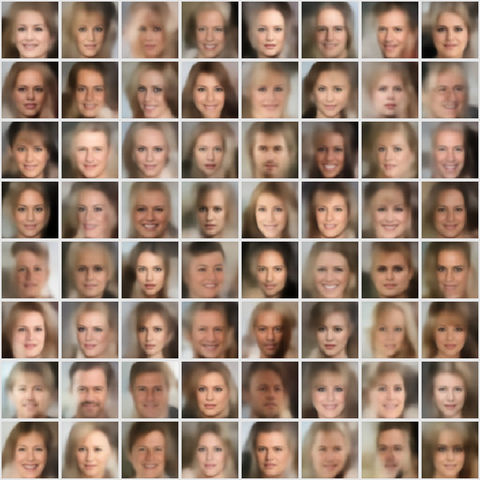}
\label{fig:aae_samples_8k_latents}
\caption{AAE samples}
}\end{subfigure}
\caption{Samples from VAE and AAE with 8k latents compared to data on CelebA.}
\label{fig:samples_8k_celeba}
\end{center}
\end{figure*}

\section{Connection to Wasserstein Autoencoders}\label{app:wasserstein_autoencoders}

\citet{tolstikhin2017wasserstein} prove the connection between using optimal transport to minimize the distance the true data distribution and the model distribution induced by an \textit{implicit} model and using autoencoders which minimize a distance between the marginal posterior and prior in latent space. Specifically, they show that:

\begin{theorem}
\label{th:wa}
Let $c$ be a measurable cost function with values in $R_+$ and $\mathbb{P}(\tdistx, \modeldist(\vx))$ the set of all joint distributions with marginals $\tdistx$ and $\modeldist(\vx)$, respectively. \\
For a model where $\modeldist(\vx|\vz)$ is a Dirac delta function, ie. $\vz$ is mapped to $\vx$ deterministically $\vx = G(\vz)$, the following holds:
\begin{align}
\label{eq:inf_dec}
 & \inf_{\Gamma \sim \mathbb{P}(\tdistx, \modeldist(\vx))} \mathbb{E}_{(\vx, \vy) \sim \Gamma} [c(\vx,\vy)] \nonumber \\
 & =
  \inf_{\vardistz: \qmarg = \zdist} \mathbb{E}_{\vx \sim \tdistx}\mathbb{E}_{\vz \sim \vardistz} [c(\vx, G(\vz))]
\end{align}
\end{theorem}

Theorem~\ref{th:wa} shows that for implicit models, encoder-decoder models can be introduced as a way to make the optimal transport computation tractable, as optimizing over the space of joint distributions $ \mathbb{P}(\tdistx, \modeldist(\vx))$ is not feasible - another approach to make the computation tractable, when $c$ is a metric, is to use the Kantorovich-Rubinstein duality \cite{wgan}.

Similarly, when doing maximum likelihood - minimizing the KL divergence between the model and data distribution - encoder-decoder models are introduced to overcome the intractability introduced by Equation~\ref{eq:marginal_x}, using Jensen's inequality (Equation~\ref{eq:free_energy}).

The connections between optimal transport and maximum likelihood do not stop here - \citet{tolstikhin2017wasserstein} optimize the RHS of Equation~\ref{eq:inf_dec} using a relaxation that adds a penalty to the objective which forces the marginal $\qmarg$ close to $\zdist$:
\begin{align}
\label{eq:constraint_wa}
\mathbb{E}_{\vx \sim \tdistx}\mathbb{E}_{\vz \sim \vardistz} [c(\vx, G(\vz))] + \lambda D_{\vz}(\qmarg, \zdist)
\end{align}

When setting $c$ to the $l_2$  or $l_1$ distance (corresponding to a Gaussian or Laplacian likelihood in the explicit model case), and setting $D_{\vz}$ to the KL divergence, we obtain the bound obtained by doing the ELBO surgery performed on the maximum likelihood objective and the training objective used to train Adversarial Autoencoders.

Hence under certain conditions, optimal transport and maximum likelihood problems lead to encoder-decoder architecture and similar optimization criteria, but allow for different modeling choices. In variational inference, the choice of $\modeldist(\vx|\vz)$ in decides the reconstruction cost function, while in Wasserstein Autoencoders $\modeldist(\vx|\vz)$ has to be a Dirac delta function, but the reconstruction cost is chosen by the practitioner. Like Adversarial Autoencoders, by \textit{approximating} $D_{\vz}$ using the density ratio trick or the Wasserstein GAN objective, Wasserstein Autoencoders lose a meaningful quantity to track - which is not the case for Variational Autoencoders which use the closed form of the KL divergence.

The connection between optimal transport and maximum likelihood opens new avenues for research that we leave for future work, while the different modeling choices provide new flexibility to machine learning practitioners.

\section{Learning with density ratios: synthetic data experiments}
\label{sec:density_ratio_gradients}

We show that the density ratio trick can be used for learning, even though it cannot be used for estimating divergences. We also show the challenges of scaling the density ratio trick to higher dimensions. To do so, devise a set of synthetic experiments where the true KL divergence is known and where we can determine how this approach scales with data dimensionality.

We use Gaussian distributions, defined by passing a random normal vector through an affine transformation with $ \myvec{W} \in \Reals^{kd}, \vb \in \Reals^d, \vz \in \Reals^k$:
\begin{equation}
\hspace{-3mm} \vz \sim \mathbb{N}(0, \mathbb{I}_k), \vx = \myvec{W}^{\top} \vz + \vb \implies x \sim \mathbb{N}(\vb, \myvec{W}^{\top} \myvec{W})
\label{eq:gauss_def}
\end{equation}
To ensure $\myvec{W}^{\top} \myvec{W}$ is full rank, we set $d = k / 10$ in all our experiments.

We first train a classifier to distinguish between two such Gaussian distributions, for varying values of $d$. In the first setting, we are concerned with \textit{divergence estimation}, keeping both distributions fixed and only learn the density ratio using the classifier. Once the classifier is trained, we report the difference between the estimated and true KL divergence values. In the second setting, we are concerned with \textit{divergence minimization} and we begin with the same initialization for the two distributions, but learn the parameters of the second Gaussian ($\myvec{W}$~and $\vb$ in Equation~\ref{eq:gauss_def}) to minimize the estimated divergence between the two distributions. This is a GAN training regime, where the generator loss is given by the reverse KL generator loss \citep{implicitgen}: $-\log \frac{D(\vx)}{1-D(\vx)}$. We track the true KL divergence during training together with the online classifier estimated divergence - we should not expect the latter to be accurate, the classifier is not trained to optimality for each update of the learned distribution, as the two models are trained jointly.

If for the same classifier that failed to approximate the true KL divergence in the estimation experiments we observe a  decrease in true divergence in the learning experiments, we can conclude that while the density ratio trick might not be a useful for estimation, it can still be used as an optimization tool. To ensure our conclusions are valid, we control over hyper-parameters, classifier architectures and random seeds of the Gaussian distributions, and average results over 10 runs.

\begin{figure*}[htb]
\centering
\includegraphics[width=\textwidth]{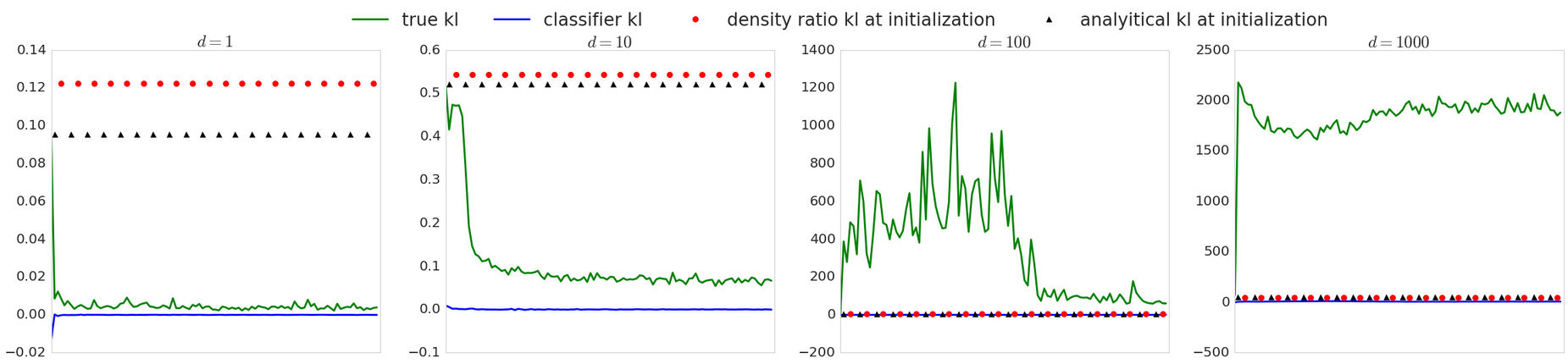}
\caption{Divergence estimation and minimization of Gaussian distributions, for different data dimensions $d$. We plot training progress using the true KL divergence between the learned and true distributions. As a reference point, we show the true KL divergence at initialization, together with how well the same classifier architecture is able to estimate the initial true KL when the two Gaussian distributions are stationary. Results are averaged over 10 different initializations for the classifier.}
\label{fig:gaussian_synth}
\end{figure*}

\begin{figure*}[htb]
\centering
\includegraphics[width=\textwidth]{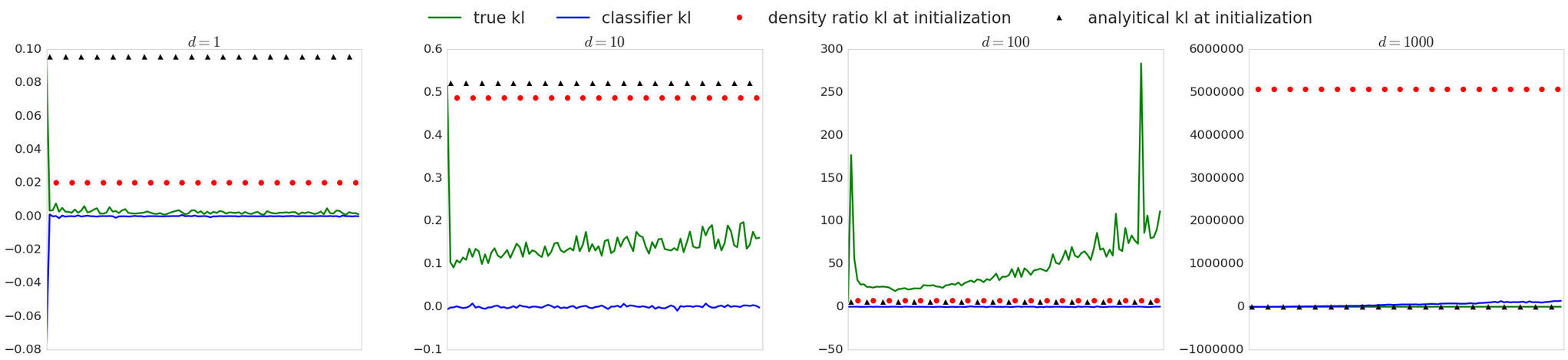}
\caption{Synthetic Gaussian experiments, for data different dimensions $d$, with a higher learning rate for the discriminator: 0.001. The model is more unstable, no longer being able to converge for data of dimensionality 100. Results are averaged over 10 different initializations for the discriminator.}
\label{fig:gaussian_synth_4_0_001}
\end{figure*}

\begin{figure*}[htb]
\centering
\includegraphics[width=\textwidth]{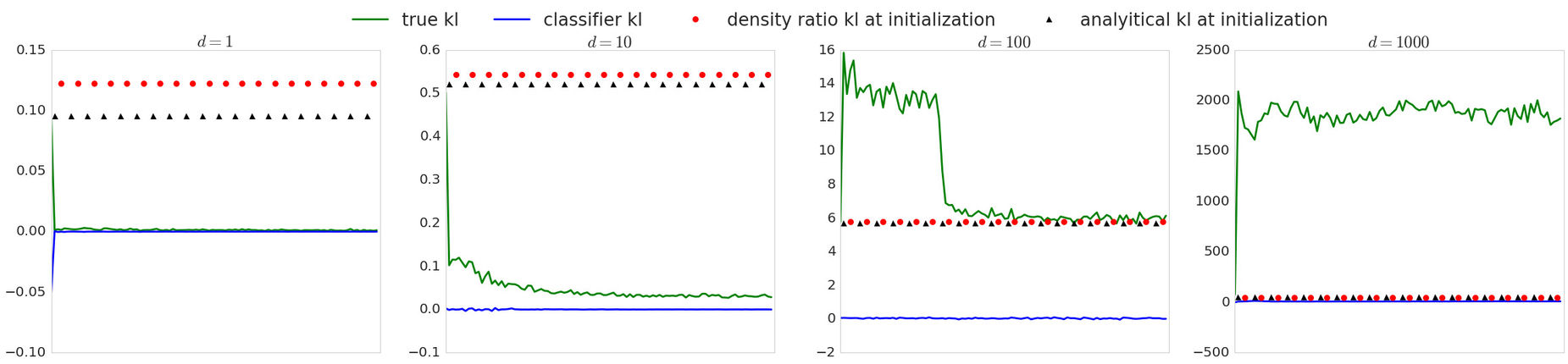}
\caption{Synthetic Gaussian experiments, for data different dimensions $d$, with 5 discriminator updates. The experimental set up is exactly the same as Figure~\ref{fig:gaussian_synth}, including random seeds and hyperparameters but the discriminator is updated 5 times per generator update. Making the discriminator updates more frequent makes the learned model converge earlier for $d=1, 10, 100$ but no improvement for $d=1000$. We see no improvement in the estimated KL, even in cases where the discriminator could estimate the KL when trained to optimality, reported as 'density ratio KL as initialization'.}
\label{fig:gaussian_synth_4_0_0001_5_updates}
\end{figure*}

Our main findings are summarized in Figures~\ref{fig:gaussian_synth},~\ref{fig:gaussian_synth_4_0_001}, and~\ref{fig:gaussian_synth_4_0_0001_5_updates} and reveal that using density ratios for learning does not reliably scale with data dimensionality. For lower dimensional data (1 and 10 dimensions), the model is able to decrease the true KL divergence. However, for higher dimensional Gaussians (dimensions 100 and 1000), a classifier with 100 million parameters (4 layer MLP) is not able to provide useful gradients and learning diverges (rightmost plot in Figure 2). Regardless of data dimensionality, the estimate of the true KL divergence provided by the density ratio trick was not accurate.

The discriminator was trained with the AdamOptimizer with $\beta_1$ set to 0.5 and $\beta_2$ set to 0.9 for 1000000 iterations. Unless otherwise specified, the discriminator was a 4 layer MLP, trained with a learning rate of $0.0001$. The learning rate used for learning the Gaussian was $0.001$. Similar results were obtained for different learning rates for the discriminator and the linear model of the Gaussian distribution.

\section{Estimating $\KLpq{\qmarg}{\zdist}$}\label{app:marginal_kl_estimation}

 We present the details of marginal KL estimation experiments described in Sections~\ref{sec:marginal_kl} and ~\ref{sec:density_ratios_latent_space}.

\subsection{Estimating $\KLpq{\qmarg}{\zdist}$ using the Monte Carlo approach}\label{app:marginal_kl_mc}

Algorithm~\ref{alg:pseudocode_mc} describes the approach used to estimate $\KLpq{\qmarg}{\zdist}$ via Monte Carlo methods. While this approach is computationally expensive, it the most accurate one. We now describe the details of this computation. For each $\vx_i$ we used $10^6$ samples from $q(\vz|\vx)$ to estimate $\log \frac{\qmarg}{p(\vz)}$. To estimate $\qmarg$ for latent sample we used the entire dataset training and validation split of the dataset at hand. In all our figures and tables, this number is reported as $N$.

\begin{algorithm}
\caption{Pseudocode for estimating the marginal KL using MC}
\label{alg:pseudocode_mc}
\begin{algorithmic}[1]
\State Load trained variational model with prior $\zdist$, and posterior $\vardistz$.
\State $\mathsf{marginal\_kl} = 0.0$
\For{$\mathsf{i}=1:\mathsf{num\_z}$}
  \State sample $\vx_{i}$ from $\tdistx$, sample $z_{i}$ from $q_{\eta}(\vz|\vx_{i})$
  \State $\mathsf{posterior\_list} = []$
  \For{$\mathsf{\vx}$ in dataset}
      \State append $\log \vardistz (\vz_{i})$ to $\mathsf{posterior\_list}$
  \EndFor
  \State $\log q(\vz_{i}) = \mathsf{log\_mean\_exp} (\mathsf{posterior\_list})$
  \State $\mathsf{marginal\_kl} += \log q(\vz_{i}) - \log p(\vz_{i})$
\EndFor
\State $\mathsf{marginal\_kl} = \mathsf{marginal\_kl} / \mathsf{num\_z}$
\end{algorithmic}
\end{algorithm}

\subsection{Estimating $\KLpq{\qmarg}{\zdist}$ using the density ratio trick}\label{app:disc_ratio_trick}

To estimate $\KLpq{\qmarg}{\zdist}$ using the density ratio trick as shown in Figure~\ref{fig:density_estimation_marginal_kl_failure} we used Algorithm~\ref{alg:pseudocode_ratio}. For all datasets, we noticed that this approach is highly sensitive to hyperparameters. We explain this two fold: first, this approach relies on the probabilities reported by a neural network classifier, which have been known to be inaccurate. New methods have been proposed to address this issue \citep{guo2017calibration}, and we leave exploring these approaches for future work. Second, as shown in Section~\ref{app:marginal_kl_density}, the distribution $q(z)$ can be very complex, making it hard for the classifier to learn to distinguish between samples from the two distributions.

We show the hyperparameter sensitivity by training different models to estimate the marginal $\KLpq{\qmarg}{\zdist}$ for the same VAE and report the different values obtained. All trained density ratio estimators were MLPs with Leaky Rectified activations of slope 0.2 and were trained for $5 * 10*5$ steps using the AdamOptimizer with $\beta_1$ and $\beta_2$ equal to 0.9.

Results on ColorMNIST  are summarized in Table~\ref{tab:kl_different_disc_cmnist}, while CelebA results are summarized in Table~\ref{tab:kl_different_disc_celeba}.

While analyzing these results, we observed that adding noise to the activation of the classifier resulted in a better classifier, but also a more confident one, which underestimates the probability that a sample was given by the prior, and the KL value being over estimated. We also see that the resulting KL is quite sensitive to the the architecture of the classifier, with an extra layer resulting in a substantial value increase for the Color MNIST case. Gradients penalties and dropout did not result in a big change in the estimated value.

\begin{algorithm}
\caption{Pseudocode for estimating the marginal KL using the density ratio trick}
\label{alg:pseudocode_ratio}
\begin{algorithmic}[1]
\State Load trained variational model with prior $\zdist$, and posterior $\vardistz$.
\State Initialize code discriminator parameters $\vomega$ randomly.
\For{$\mathsf{iter}=1:\mathsf{max\_iter}$}
  \State Update parameters $\vomega$ by maximizing   $\E_{\tdistx} \E_{\vardistz} \bigl[ \log(\codediscz)\bigr] + \E_{\zdist} \bigl[\log(1 - \codediscz)\bigr]$
\EndFor
\State $\mathsf{marginal\_kl} = 0.0$
\For{$\mathsf{i}=1:\mathsf{num\_z}$}
  \State sample $\vx_{i}$ from $\tdistx$, sample $\vz_{i}$ from $q_{\eta}(\vz|\vx_{i})$
  \State $\mathsf{marginal\_kl} +\!\!= \log\codedisc(\vz_{i}) - \log (1 - \codedisc(\vz_{i}))$
\EndFor
\State $\mathsf{marginal\_kl} = \mathsf{marginal\_kl} / \mathsf{num\_z}$
\end{algorithmic}
\end{algorithm}

\begin{table}[htpb]
\small
\centering
\caption{\label{tab:kl_different_disc_cmnist} Estimating a marginal KL using the density ratio trick for a standard VAE with 50 latents trained on Color MNIST. The number of hidden units per layer was 5000. When the KL is estimated numerically, the result is $12.3$. From Equation~\eqref{eq:mi} and that the mutual information term is bound by $\log N$, with $N=60000$ and the average posterior KL of the model is $23.34$, we know that the value needs to be greater than $12.46$. All models used a learning rate of $0.0005$.}
\vspace{0.2cm}
\begin{tabular}{p{1cm}|p{1.2cm}|p{1.3cm}|l}
\# layers  & Gradient Penalty & Activation Noise & KL \\ \midrule
$3$ & No & No & 2.3 \\
$4$ & No &  No & 3.3 \\
$4$ & No &  Yes & 25812.8 \\
$4$ & Yes &  No & 3.1 \\
\end{tabular}
\end{table}

\begin{table}[htpb]
\small
\centering
\caption{\label{tab:kl_different_disc_celeba} Estimating a marginal KL using the density ratio trick for a standard VAE with 100 latents trained on CelebA. The number of hidden units per layer was 5000. When the KL is estimated numerically, the result is $100.3$. From Equation~\eqref{eq:mi} and that the mutual information term is bound by $\log N$, with $N=162770$ and the average posterior KL of the model is $112.37$, we know that the value needs to be greater than $100.0$.}
\vspace{0.2cm}
\begin{tabular}{p{1cm}|p{1.2cm}|p{1.3cm}|c|c|c}
\# layers & Learning Rate &  Activation Noise & Dropout & KL \\ \midrule
$5$ & 0.0005 & No  & No & 17.7 \\
$5$ & 0.0005 & Yes &  No & 720140.9 \\
$5$ & 0.0001 & No  &  No & 14.8 \\
$7$ & 0.0005 & No  &  No & 18.8 \\
$7$ & 0.0005 & No  &  Yes & 19.0 \\
$7$ & 0.0001 & No  &  No & 18.0 \\
$10$ & 0.0001& No  &  No & 18.1 \\
\end{tabular}
\end{table}

\subsection{Estimating $\KLpq{\qmarg}{\zdist}$ using a density model for $q(\vz)$}\label{app:marginal_kl_density}

\begin{figure*}[hb]
\centering
\begin{subfigure}{0.45\textwidth}{
\includegraphics[height=3.1cm]{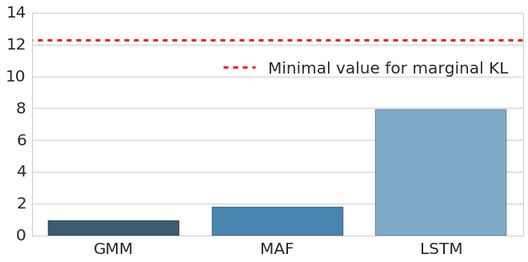}
\caption{ColorMNIST}
}\end{subfigure}
\hspace{3mm}
\begin{subfigure}{0.45\textwidth}{
\includegraphics[height=3.1cm]{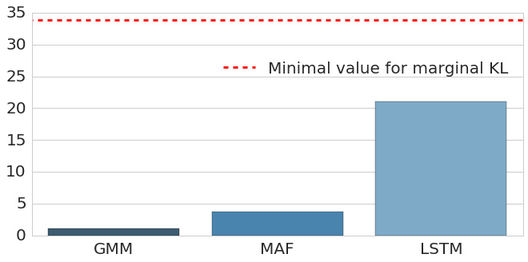}
}\end{subfigure}
\caption{Estimating $\KLpq{\qmarg}{\zdist}$ using different density models to estimate $q(z)$: Gaussian Mixture Models (GMM), Masked Auto-regressive Flow (MAF) and autoregressive models (LSTM). We plot the minimal value for the marginal KL - computed from Equation~\ref{eq:mi} - which allows us to conclude that all three density estimation approaches underestimate the true KL. LSTMs outperform the other models, showing that the autoregressivity of these models is necessary to model $\qmarg$.}
\label{fig:marginal_kl_estimation_density_models}
\end{figure*}

We compare the MonteCarlo and density ratio trick approach with a third way to estimate $\KLpq{\qmarg}{\zdist}$: by using a density model to learn  $q(\vz)$. We use three density models: Gaussian Mixture Model, Masked Auto-regressive Flows \citep{maf}, and Gaussian auto-regressive models implemented using an \citep{lstm}. Algorithm~\ref{alg:pseudocode_q_z} was used to learn these models.
Across models, we show a failure  (see Figure~\ref{fig:marginal_kl_estimation_density_models}) to reach the theoretical bound, showing that $\qmarg$ is a complex distribution.

\mysubsection{Diagonal Gaussian Mixture Models}\\
In this setting, we model $\qmarg$ using:
\begin{equation*}
 \qmarg = \sum_i^{k} \pi_i \ \mathcal{N}(\vz| \myvec{\mu}_i, \myvec{\sigma}_i), \sum_i^{k} \pi_i = 1
\end{equation*}

\mysubsection{Masked Auto-regressive Flows}\\
In this setting, the density model is given by transforming a standard Gaussian using auto-regressive models as normalizing flows:

\begin{equation*}
 \qmarg = \mathcal{N}(0, \mathbb{I} | \vz) \left\lvert \det \left(\frac{d\ f^{-1}}{dz} \right) \right\rvert
\end{equation*}

where $f$ has to be an invertible function for which the determinant of its Jacobian is easy to compute. In practice, we leverage the fact that the composition of two functions which have these proprieties also has this property to chain a number of transforms.

\mysubsection{Gaussian Auto-regressive models}\\
The auto-regressivity of the recurrent neural network was used to model $q(z_i| z_{<i})$:
\begin{equation*}
 \qmarg = \prod q(z_i| z_{<i}) = \prod \mathcal{N}(z_i | \mu(z_{<i}, \sigma(z_{<i})))
\end{equation*}

\newpage %
\begin{algorithm}[htbp]
\caption{Pseudocode for estimating the marginal KL using a density estimator for $q(\vz$)}
\label{alg:pseudocode_q_z}
\begin{algorithmic}[1]
\State Load trained variational model with prior $\zdist$, and posterior $\vardistz$.
\State Initialize density $\vt$ model parameters $\vomega$ randomly.
\For{$\mathsf{iter}=1:\mathsf{max\_iter}$}
  \State Update parameters $\vomega$ by maximizing    $\E_{\tdistx} \E_{\vardistz} \bigl[ \log(\vt(\vz))\bigr]$
\EndFor
\State $\mathsf{marginal\_kl} = 0.0$
\For{$\mathsf{i}=1:\mathsf{num\_z}$}
  \State sample $\vx_{i}$ from $\tdistx$, sample $\vz_{i}$ from $q_{\eta}(\vz|\vx_{i})$
  \State $\mathsf{marginal\_kl} += \log(\vt(\vz_{i}) - \log p(\vz_{i})$
\EndFor
\State $\mathsf{marginal\_kl} = \mathsf{marginal\_kl} / \mathsf{num\_z}$
\end{algorithmic}
\end{algorithm}

\section{\ourgan~loss function and pseudocode}\label{app:our_gan_pseudocode}

In all the equations below, $D_{\theta}$ refers to the data discriminator, while $\codedisc$ refers to the code discriminator. They are both trained to minimize a cross entropy loss, like in the original GAN formulation.

The loss function for \reconourgan~is:
\begin{align*}
& \E_{\vardistz} \Big[ -\lambda||\vx-\generator(\vz)||_1 + \log \frac{\disc(\generator(\vz))}{1-\disc(\generator(\vz))} \nonumber \\
& + \log \frac{\codediscz}{1-\codediscz}\Big]
\end{align*}

In contrast, the loss function for \ourgan~is:
\begin{align*}
& \E_{\vardistz}  \Big[-\lambda||\vx-\generator(\vz)||_1 + \log \frac{\disc(\generator(\vz))}{1-\disc(\generator(\vz))} \\
& + \log \frac{\codediscz}{1-\codediscz} \Big] + \E_{p(\vz)} \log \frac{\disc(\generator(\vz))}{1-\disc(\generator(\vz))}
\label{eq:our_gan_loss_recon)}
\end{align*}

The overall training procedure is summarized in Algorithm~\ref{alg:pseudocode}.
\begin{algorithm*}
\small
\caption{Pseudocode for \ourgan}
\label{alg:pseudocode}
\begin{algorithmic}[1]
\State Initialize parameters of generator $\vtheta$, encoder $\veta$, discriminator $\vphi$ and code discriminator $\vomega$ randomly.
\State Let $\vzhat\sim\vardistz$ denote a sample from $\vardistz$ and $\vxhat = \generator(\vzhat)$ denote the `reconstruction' of $\vx$ using $\vzhat$.
\State Let $R_{\disc}(\vx) = - \log{\disc(\vx)} + \log{(1- \disc(\vx))}$
\State Let $R_{\codedisc}(\vz) = - \log{\codedisc(\vz)} + \log{(1- \codedisc(\vz))}$
\For{$\mathsf{iter}=1:\mathsf{max\_iter}$}
\State Update encoder $\veta$ by minimizing
\newline\indent\algcomment{ data reconstruction and code generation loss}
\begin{align}
& \E_{\tdistx}  \E_{\vardistz} \Big[ \lambda||\vx-\generator(\vz)||_1 \bigr]  + R_{\codedisc}({\vz})  \Bigr] \approx \E_{\tdistx} \Big[ \lambda||\vx-\vxhat||_1 \bigr] + R_{\codedisc}({\vzhat})\Bigr]
\end{align}

\State Update generator $\vtheta$ by minimizing
\newline\indent\algcomment{data reconstruction and generation loss}
\begin{align}
& \mathbb{E}_{\tdistx}\E_{\vardistz} \bigl[ \lambda||\vx-\generator(\vz)||_1  + R_{kl}(\vz)\bigr] \nonumber + \E_{\zdist} \bigl[ R_{\disc}(\generator(\vz))\bigr] \\
& \approx \mathbb{E}_{\tdistx}  \bigl[ \lambda||\vx-\vxhat||_1 + R_{\disc}(\vxhat) \bigr] \nonumber + \E_{\zdist} \bigl[ R_{\disc}(\generator(\vz)) \bigr]
\end{align}

\State Update discriminator $\vphi$ by minimizing
\newline\indent \algcomment{treat data as real, reconstructions and samples as fake}
\begin{align}
& \mathbb{E}_{\tdistx}\bigl[-2 \log\discx  - \E_{\vardistz}\log\bigl(1-\disc(\generator(\vz))\bigr)
\bigr] \nonumber + \E_{\zdist} \bigl[- \log\bigl(1-\disc(\generator(\vz))\bigr)\bigr] \\
&\approx \mathbb{E}_{\tdistx}\bigl[-\log\discx  - \log\bigl(1-\disc(\vxhat)\bigr)
\bigr] \nonumber + \E_{\zdist} \bigl[- \log\bigl(1-\disc(\generator(\vz))\bigr)\bigr]
\end{align}

\State Update code discriminator $\vomega$ by minimizing
\newline\indent\algcomment{treat $p(\vz)$ as real and codes from the encoder as fake}
\begin{align}
& \E_{\tdistx} \E_{\vardistz} \bigl[ -\log(1-\codediscz)\bigr] + \E_{\zdist} \bigl[-\log\codediscz\bigr]
\approx  \E_{\tdistx}  \bigl[-\log(1-\codedisc(\vzhat))\bigr] + \E_{\zdist} \bigl[ -\log(\codediscz)\bigr]
\end{align}
\EndFor
\end{algorithmic}
\end{algorithm*}

\section{The effect of the visible distribution in VAE training}\label{app:qn_bernoulli}

\begin{figure*}[htbp]
\begin{center}
\begin{subfigure}{0.9\textwidth}{
\centering
\includegraphics[width=0.3\textwidth]{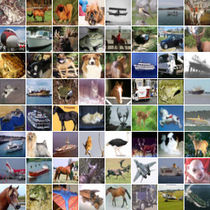} \addhspace
\includegraphics[width=0.3\textwidth]{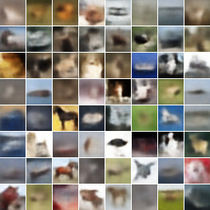} \addhspace
\includegraphics[width=0.3\textwidth]{cifar_vae_samples_last} \addhspace
\caption{Bernoulli}
\label{fig:cifar_bern}
}\end{subfigure}
\begin{subfigure}{0.9\textwidth}{
\centering
\includegraphics[width=0.3\textwidth]{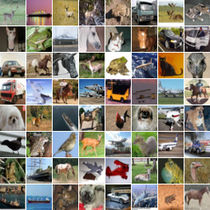} \addhspace
\includegraphics[width=0.3\textwidth]{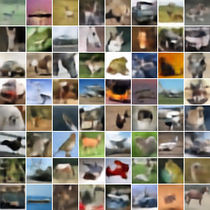} \addhspace
\includegraphics[width=0.3\textwidth]{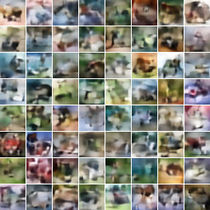} \addhspace
\caption{QuantizedNormal}
\label{fig:cifar_qn}
}\end{subfigure}
\caption{Comparisons of reconstructions and samples generated from VAEs using Bernoulli and Quantized Normal pixel distributions. The observed trade-off between reconstruction and sample quality is consistent throughout different hyperparameters. For the models displayed here, the difference can be seen in the different KL values obtained in the loss function used to train the models: 44.7 for the Bernoulli model,  and 256.7 for the QuantizedNormal model.}
\label{fig:cifar:qn_bernoulli_recon}
\end{center}
\end{figure*}

Figure~\ref{fig:cifar:qn_bernoulli_recon} visually shows the trade-off seen between using a Bernoulli or a QuantizedNormal distribution as the visible pixel distribution, $p(\vx|\vz)$ in VAEs.

We now unpack the mathematical justification for why the Bernoulli distribution produces worse reconstructions. We will perform the analysis for a pixel $x$, but this straightforwardly extends to entire images. Assume a Bernoulli distribution with mean $\mu \in [0, 1]$. Then the Bernoulli loss is $x\mu - \log (1 + e ^{\mu})$. The gradient of the loss is $x - \sigma(\mu)$, where $\sigma$ is the sigmoid function. If we use a Gaussian distribution with mean $m$ and standard deviation $s$, the gradient is $\frac {x - m}{s^2}$. We can see that the gradients of the two distributions have the same form, and using a Bernoulli distribution is equivalent to using a Gaussian distribution with variance 1. By setting the variance to 1, using a Bernoulli likelihoods spreads mass around each pixel and cannot specialize to produce good reconstructions.

\section{Tracking the variational lower bound - VEEGAN}\label{app:veegan_bound_tracking}

\begin{figure*}[htpb]
\begin{center}
\begin{subfigure}{0.4\textwidth}{
\includegraphics[width=\textwidth]{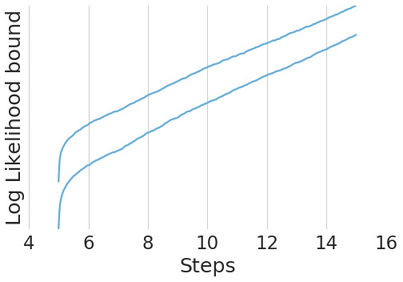}
\caption{VAE}
}\end{subfigure}
\begin{subfigure}{0.4\textwidth}{
\includegraphics[width=\textwidth]{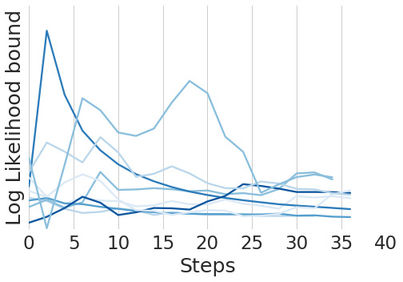}
\caption{VEEGAN}
}\end{subfigure}
\caption{
\label{fig:bound_tracking}
  CIFAR-10 training curves of a standard VAE and VEEGAN across different hyperparameters. Results were obtained on . Since the bound of VEEGAN is not obtained on the observed data the numbers are not directly comparable. The aim of this plot is to show the \textbf{trend} of training, as we expect that as model training progresses, the likelihood increases. We see that for VEEGAN this is not the case, even though the models perform comparable with state of the art (Figure~\ref{fig:mssim_inception}).}
\end{center}
\end{figure*}

In this section we compare the evidence lower bound training behavior, between VAEs and VEEGAN. VEEGAN introduces a new bound,
\begin{equation}
 \KLpq{\modeldist(\vx|\vz)p(\vz)}{p_{\gamma}(\vz|\vx)p(\vx)} - \mathbb{E} [ \log(p(\vz))] + l_1(\vz, F_{\theta}(\vx))
\label{eq:veegan_bound}
\end{equation}

where $F_{\theta}$ is the reconstructor network used to build the implicit $p(\vx|\vz)$ and the KL divergence is estimated using the density ratio trick. The expected and desired behavior is that the bound increases as training progresses, however, as seen in Figure~\ref{fig:bound_tracking}, variational hybrids do not solve one of the fundamental problems with adversarial model training, namely introducing a quantity to use to assess convergence.

\vspace{-2mm}
\section{Real data evaluation metrics}\label{sec:metrics}
\vspace{-2mm}

A universal metric that can assess both overfitting, sample quality and sample diversity has not been found for generative models. Instead, multiple metrics which assess different aspects of a model have been proposed. To get a better overview of model performance, we use metrics which each capture a different aspect of training.

\evalsubsection{Inception score - sample quality and between class sample diversity}
 The most popular evaluation metric for implicit models is the Inception Score \citep{improvedgan}.
 The Inception Score correlates with human sample evaluation and measures sample quality, between class sample diversity, but cannot capture withing class mode collapse (for example, the model could generate the same horse again and again, and the Inception Score will not penalize it) or overfitting. The Inception score uses the last layer logits of a ImageNet trained Inception network \citep{inception} to determine how classifiable a sample is. For generative models trained on CIFAR10, we complement our reporting by using a VGG style convolutional neural network, trained on CIFAR10, which obtained 5.5\% error.

\evalsubsection{Multi-scale structural similarity (MS-SSIM): sample diversity}
To measure sample diversity, we use 1.0 - MS-SSIM \citep{wang2003multiscale}, an image similarity metric ranging between 0.0 (low similarity) and 1.0 (high similarity) that has been shown to correlate well with human judgment. The use of MS-SSIM for sample diversity was introduced by \citet{acgan}, which used it compute in class sample similarity for conditional models, as between class variability can lead to ambiguous results. For models trained on the CelebA dataset, we can use this sample diversity metric, since the dataset only contains faces.

\evalsubsection{Independent Wasserstein critic - sample quality and overfitting} \citet{ivogan} and \citet{jiwoong2018quantitatively} proposed training an independent Wasserstein GAN critic to distinguish between real data and generated samples. This metric has been shown to correlate with human evaluations \citep{jiwoong2018quantitatively}, and if the independent critic is trained on validation data, it can also be used to measure overfitting \citep{ivogan}.
All our reported results using the Independent Wasserstein Critic use a WGAN-GP model, trained to distinguish between the data validation set and model samples.

\section{AdversarialVB results}
\label{app:adv_vb}
\begin{figure*}[htbp]
\begin{center}
\includegraphics[width=0.42\textwidth]{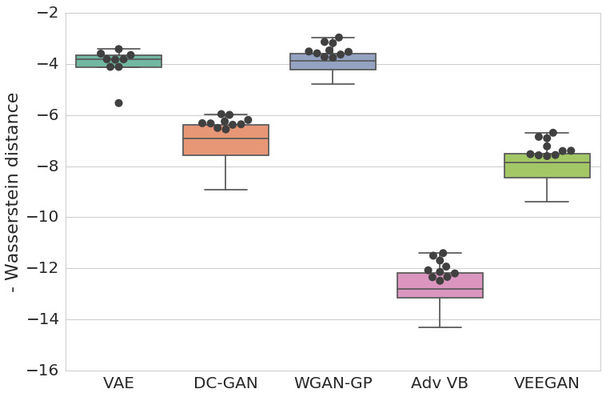} \addhspace
\includegraphics[width=0.28\textwidth]{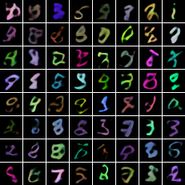}
\caption{Results obtained using Adversarial VB, without adaptive contrast. The model was not able to match $\qmarg$ and $p(\vz)$, and this results in an independent critic being able to easily distinguish samples from data (left). This can be seen visually on the right, as the digits do not appear well defined for a large number of samples.}
\label{fig:adv_vb}
\end{center}
\end{figure*}

Results obtained using AdversarialVB are presented in Figure~\ref{fig:adv_vb}.

\section{Training details: hyperparameters and network architectures\label{app:hyperparameters}}

 For all our models, we kept a fixed learning rate throughout training. We note the difference with AGE, where the authors decayed the learning rate during training, and changed the loss coefficients during training\footnote{As per advice found here: \url{https://github.com/DmitryUlyanov/AGE/}}.). The exact learning rate sweeps are defined in Table ~\ref{tab:learning_rates}. We used the Adam optimizer \citep{adam} with $\beta_1 = 0.5$ and $\beta_2 = 0.9$ and a batch size of 64 for all our experiments. We used batch normalization \citep{batchnorm} for all our experiments. We trained all \colormnist~ models for 100000 iterations, and CelebA and CIFAR-10 models for 200000 iterations.

\begin{table*}[h]
\centering
\resizebox{1.\textwidth}{!}{
\begin{tabular}{@{}c|c|c|c|c|c@{}}
Network  & VAE &  \specialcell{DCGAN \\WGAN-GP} & AAE & \specialcell{\reconourgan~\\ \ourgan} & VEEGAN \\ \midrule
Generator/Encoder & $0.001, 0.0005, 0.005$ &  $0.0001, 0.0002, 0.0003$ & $0.001, 0.0005, 0.005$ &  $0.0001, 0.0005$ & $0.001, 0.0005$  \\
Discriminator  &  \NA  &  $0.0001, 0.0002, 0.0003$ & \NA & $0.0005$ &  $0.00005, 0.0001$ \\
Code discriminator& \NA  & \NA & $0.0005, 0.00005, 0.00001$ & $0.0005$  &  \NA \\
\end{tabular}
}
\vspace{0.2cm}
\caption{\label{tab:learning_rates} Learning rate sweeps performed for each model.}
\end{table*}

\mysubsection{Scaling coefficients}\\
We used the following sweeps for the models which have combined losses with different coefficients (for all our baselines, we took the sweep ranges from the original papers):
\begin{itemize}
\item WGAN-GP
  \begin{itemize}
    \item The gradient penalty of the discriminator loss function: 10.
  \end{itemize}

\item \ourgan~ and \reconourgan
  \begin{itemize}
    \item Data reconstruction loss for the encoder: sweep over 1, 5, 10, 50.
    \item Data reconstruction loss for the generator: sweep over 1, 5, 10, 50.
    \item Adversarial loss for the generator (coming from the data discriminator): 1.0.
    \item Adversarial loss for the encoder (coming from the code discriminator): 1.0.
  \end{itemize}
\end{itemize}

For Adversarial Autoencoders and VEEGAN, we followed the advice from the original paper and did not weight the different loss terms using coefficients.

\mysubsection{Choice of loss functions} \\
  For VEEGAN, we used the $l_1$ loss as the code reconstruction loss. For \reconourgan~and \ourgan~, we used $l_1$ as the data reconstruction loss and the classifier GAN loss for the data and code discriminator.

\mysubsection{Updates} \\
  For the WGAN-GP experiments, we did 5 discriminator updates for generator update. All other models used the same number updates for model component (discriminator, generator, encoder, decoder).

\mysubsection{Choice of latent prior} \\
  We use a univariate normal prior for all models.

\subsection{Network architectures}
For all our baselines, we used the same discriminator and generator architectures, and we controlled the number of latents for a fair comparison. For methods which needs an encoder such as VAEs, VEEGAN, \reconourgan~and \ourgan~, the encoder is always set as a convolutional network, formed by transposing the generator (we do not use any activation function after the encoder). All discriminators use leaky units \citep{leakyrelu} with a slope of 0.2, and all generators used ReLUs. In all VAE results, unless otherwise specified, we used a Bernoulli visibile distribution and a Gaussian latent posterior.

\mysubsection{\colormnist~} \\
For all our models trained on \colormnist, we swept over the latent sizes 10, 50 and 75. Tables~\ref{tab:color_mnist_disc} and \ref{tab:color_mnist_gen} describe the discriminator and generator architectures respectively.

\mysubsection{CelebA and CIFAR-10} \\
The discriminator and generator architectures used for CelebA and CIFAR-10 were the same as the ones used by \citet{improvedwgan} for WGAN, using code at \url{https://github.com/martinarjovsky/WassersteinGAN/blob/master/models/dcgan.py}.
Note that the WGAN-GP paper reports Inception Scores computed on a different architecture, using 101-Resnet blocks. For VEEGAN, we designed a code discriminator as defined in Table ~\ref{tab:veegan_code_disc}.

\mysubsection{Code discriminator architectures} \\
 For a fair comparison between models, we used the same code discriminator architecture, where one is applicable. We tried both deeper convolutional architectures as well as shallow but bigger linear layers. We found the latter to work best and hence we used a 3 layer MLP with 1000 units each and leaky RELUs activations \citep{leakyrelu} with a slope of 0.2 as the code discriminator. Using 5000 units did not substantially improve results. This can be explained by the fact that too strong gradients from the code discriminator can effect the reconstruction ability of the encoder, and then more careful tuning of loss coefficient is needed. Perhaps optimization algorithms which are better suited for multi loss objective could help this issue.

\begin{table}[h]
\centering
\begin{tabular}{@{}rlll@{}} \toprule
Operation              & Kernel       & Strides      & Feature maps  \\ \midrule
Convolution            & $5 \times 5$ & $2 \times 2$ & $8$   \\
Convolution            & $5 \times 5$ & $1 \times 1$ & $16$   \\
Convolution            & $5 \times 5$ & $2 \times 2$ & $32$    \\
Convolution            & $5 \times 5$ & $1 \times 1$ & $64$    \\
Convolution            & $5 \times 5$ & $2 \times 2$ & $64$    \\
Linear adv            & \NA          & \NA          & $2$     \\
Linear class     & \NA          & \NA          & $10$         \\
\end{tabular}
\vspace{0.2cm}
\caption{\label{tab:color_mnist_disc} \colormnist~data discriminator architecture used for all models which require one. For DCGAN, we use dropout of 0.8 after the last convolutional layer. No other model uses dropout.}
\end{table}

\begin{table}[h]
\vspace{0.5cm}
\centering
\resizebox{1.\columnwidth}{!}{
\begin{tabular}{@{}rlll@{}} \toprule
Operation              & Kernel       & Strides      & Feature maps    \\ \midrule
Linear                 & \NA          & \NA          & $3136$       \\
Transposed Convolution & $5 \times 5$ & $2 \times 2$ & $64$          \\
Transposed Convolution & $5 \times 5$ & $1 \times 1$ & $32$          \\
Transposed Convolution & $5 \times 5$ & $2 \times 2$ & $3$           \\
\end{tabular}
}
\caption{\label{tab:color_mnist_gen} \colormnist~generator architecture. This architecture was used for all compared models.}
\end{table}

\begin{table}[h]
\vspace{0.5cm}
\centering
\begin{tabular}{@{}rlll@{}} \toprule
Operation              & Kernel       & Strides      & Feature maps  \\ \midrule
Convolution            & $3 \times 3$ & $1 \times 1$ & $[512, 1024]$   \\
Convolution            & $2 \times 2$ & $1 \times 1$ & $[512, 1024]$   \\
Linear adv             & \NA          & \NA          & $2$         \\
\end{tabular}
\caption{\label{tab:veegan_code_disc} The joint discriminator head used for VEEGAN. The input of this network is a vector concatenation of data and code features, each obtained by passing the data and codes through the same discriminator and code architectures used for the other models (excluding the classification layer).}
\end{table}

\section{Reconstructions \label{app:reconstructions}}
\begin{figure*}[htbp]
\begin{center}
\begin{subfigure}{0.9\textwidth}{
\centering
\includegraphics[width=0.45\textwidth]{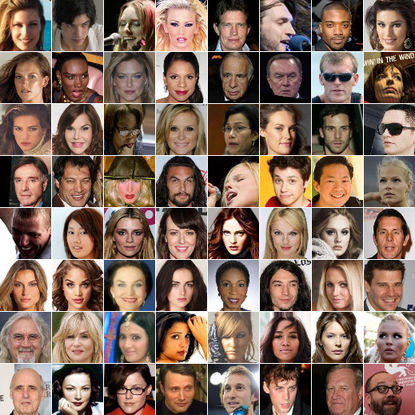} \addhspace
\includegraphics[width=0.45\textwidth]{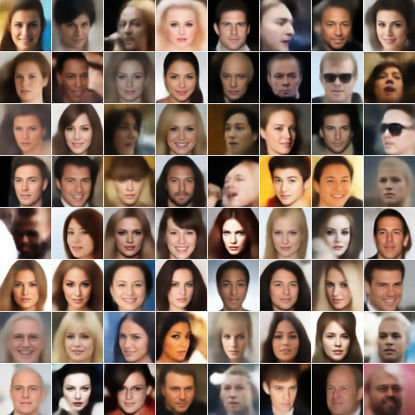} \addhspace
\caption{VAE}
\label{fig:celeba:reconstructions:vae}
}\end{subfigure}
\begin{subfigure}{0.9\textwidth}{
\centering
\includegraphics[width=0.45\textwidth]{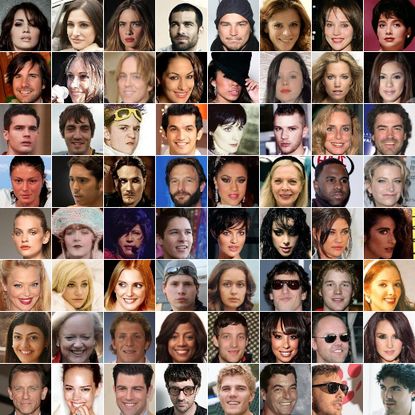} \addhspace
\includegraphics[width=0.45\textwidth]{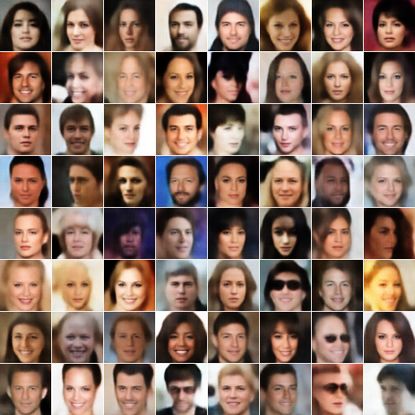} \addhspace
\caption{AAE}
\label{fig:celeba:reconstructions:advae}
}\end{subfigure}
\begin{subfigure}{0.9\textwidth}{
\centering
\includegraphics[width=0.45\textwidth]{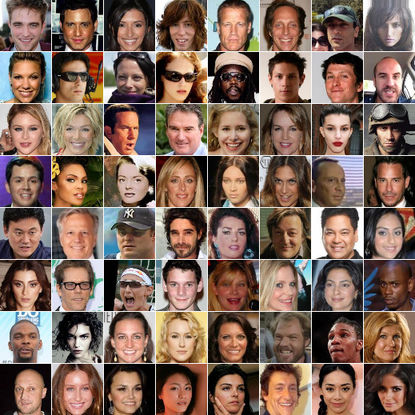} \addhspace
\includegraphics[width=0.45\textwidth]{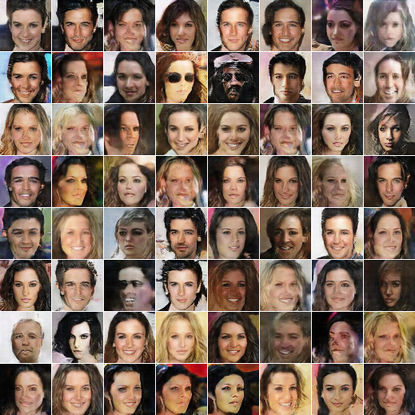} \addhspace
\caption{\ourgan}
\label{fig:celeba:reconstructions:ours}
}\end{subfigure}
\caption{
Training reconstructions obtained using a standard VAE, Adversarial Autoencoders and \ourgan~on CelebA. Left is the data and right are reconstructions.
}
\label{fig:celeba:reconstructions}
\end{center}
\end{figure*}

\begin{figure*}[htbp]
\begin{center}
\begin{subfigure}{0.9\textwidth}{
\centering
\includegraphics[width=0.4\textwidth]{cifar_vae_data_last} \addhspace
\includegraphics[width=0.4\textwidth]{cifar_vae_recons_last} \addhspace
\caption{VAE}
\label{fig:cifar:reconstructions:vae}
}\end{subfigure}
\begin{subfigure}{0.9\textwidth}{
\centering
\includegraphics[width=0.4\textwidth]{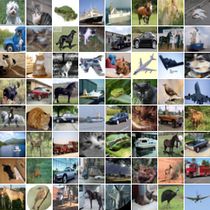} \addhspace
\includegraphics[width=0.4\textwidth]{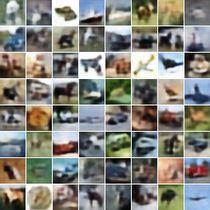} \addhspace
\caption{Adversarial Autoencoders}
\label{fig:cifar:reconstructions:advae}
}\end{subfigure}
\begin{subfigure}{0.9\textwidth}{
\centering
\includegraphics[width=0.4\textwidth]{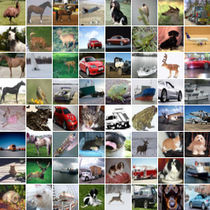} \addhspace
\includegraphics[width=0.4\textwidth]{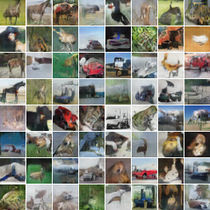} \addhspace
\caption{\ourgan}
\label{fig:cifar:reconstructions:ours}
}\end{subfigure}
\caption{
Training reconstructions obtained using a standard VAE, Adversarial Autoencoders and \ourgan~on CIFAR-10. Left is the data and right are reconstructions.
}
\label{fig:cifar:reconstructions}
\end{center}
\end{figure*}

We show reconstructions obtained using \ourgan~and VAEs for the CelebA dataset in Figure \ref{fig:celeba:reconstructions} and on CIFAR-10 in Figure \ref{fig:cifar:reconstructions}.

\section{Model samples for real data experiments \label{app:samples}}

\begin{figure*}[htbp]
\begin{center}
\begin{subfigure}{0.45\textwidth}{
\centering
\includegraphics[width=0.9\textwidth]{celeba_vae_samples_last} \addhspace
\caption{VAE}
}\end{subfigure}
\begin{subfigure}{0.45\textwidth}{
\centering
\includegraphics[width=0.9\textwidth]{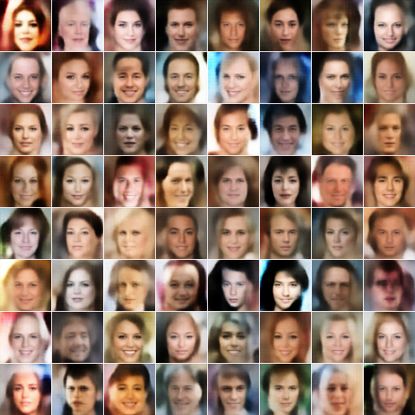} \addhspace
\caption{AAE}
}\end{subfigure}
\begin{subfigure}{0.45\textwidth}{
\centering
\includegraphics[width=0.9\textwidth]{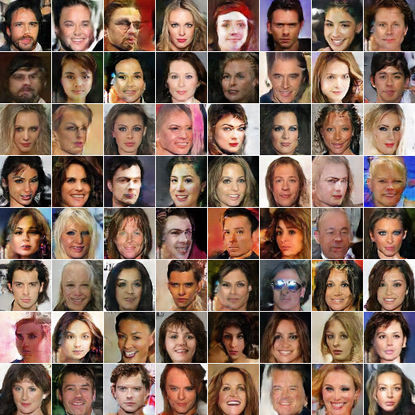} \addhspace
\caption{DCGAN}
}\end{subfigure}
\begin{subfigure}{0.45\textwidth}{
\centering
\includegraphics[width=0.9\textwidth]{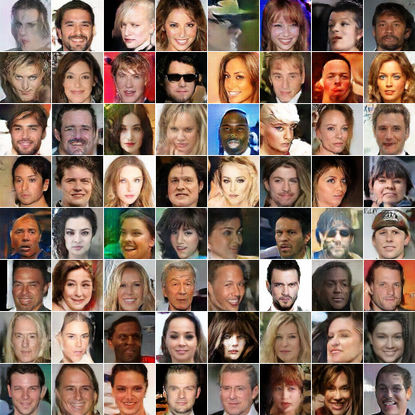} \addhspace
\caption{WGAN-GP}
}\end{subfigure}
\begin{subfigure}{0.45\textwidth}{
\centering
\includegraphics[width=0.9\textwidth]{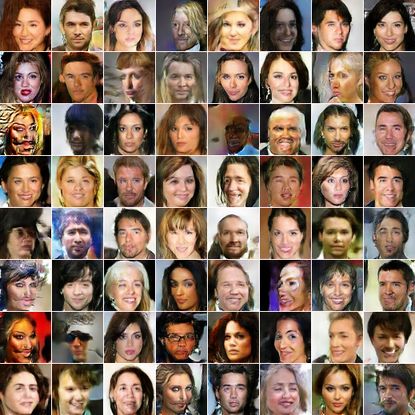} \addhspace
\caption{VEEGAN}
}\end{subfigure}
\begin{subfigure}{0.45\textwidth}{
\centering
\includegraphics[width=0.9\textwidth]{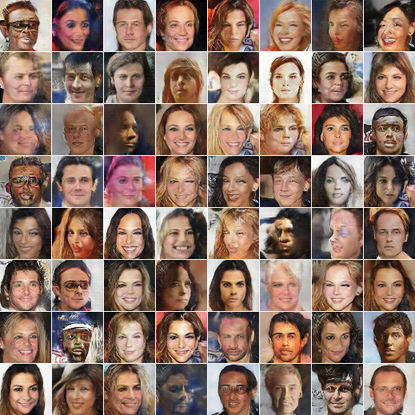} \addhspace
\caption{\ourgan}
}\end{subfigure}
\caption{CelebA samples.}
\label{fig:celeba:samples}
\end{center}
\end{figure*}

\begin{figure*}[htbp]
\begin{center}
\begin{subfigure}{0.45\textwidth}{
\centering
\includegraphics[width=0.8\textwidth]{cifar_vae_samples_last} \addhspace
\caption{VAE}
}\end{subfigure}
\begin{subfigure}{0.45\textwidth}{
\centering
\includegraphics[width=0.8\textwidth]{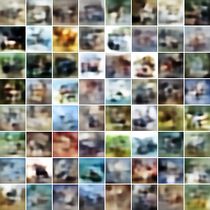} \addhspace
\caption{AAE}
}\end{subfigure}
\begin{subfigure}{0.45\textwidth}{
\centering
\includegraphics[width=0.8\textwidth]{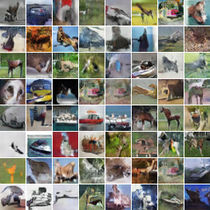} \addhspace
\caption{DCGAN}
}\end{subfigure}
\begin{subfigure}{0.45\textwidth}{
\centering
\includegraphics[width=0.8\textwidth]{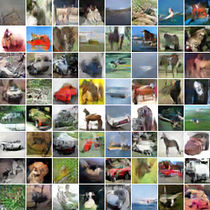} \addhspace
\caption{WGAN-GP}
}\end{subfigure}
\begin{subfigure}{0.45\textwidth}{
\centering
\includegraphics[width=0.8\textwidth]{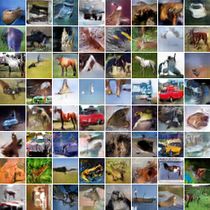} \addhspace
\caption{VEEGAN}
}\end{subfigure}
\begin{subfigure}{0.45\textwidth}{
\centering
\includegraphics[width=0.8\textwidth]{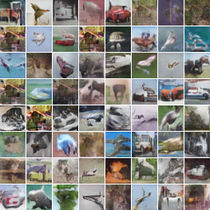} \addhspace
\caption{\ourgan}
}\end{subfigure}
\caption{CIFAR10 samples.}
\label{fig:cifar:samples}
\end{center}
\end{figure*}

\begin{figure*}[htbp]
\begin{center}
\begin{subfigure}{0.45\textwidth}{
\centering
\includegraphics[width=0.8\textwidth]{cmnist_vae_samples_last} \addhspace
\caption{VAE}
}\end{subfigure}
\begin{subfigure}{0.45\textwidth}{
\centering
\includegraphics[width=0.8\textwidth]{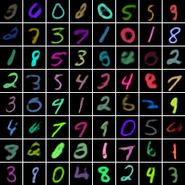} \addhspace
\caption{AAE}
}\end{subfigure}
\begin{subfigure}{0.45\textwidth}{
\centering
\includegraphics[width=0.8\textwidth]{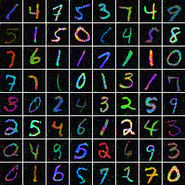} \addhspace
\caption{DCGAN}
}\end{subfigure}
\begin{subfigure}{0.45\textwidth}{
\centering
\includegraphics[width=0.8\textwidth]{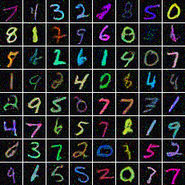} \addhspace
\caption{WGAN-GP}
}\end{subfigure}
\begin{subfigure}{0.45\textwidth}{
\centering
\includegraphics[width=0.8\textwidth]{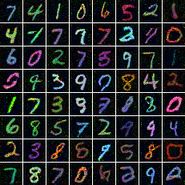} \addhspace
\caption{VEEGAN}
}\end{subfigure}
\begin{subfigure}{0.45\textwidth}{
\centering
\includegraphics[width=0.8\textwidth]{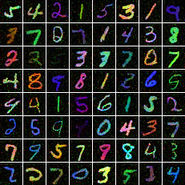} \addhspace
\caption{\ourgan}
}\end{subfigure}
\caption{ColorMNIST samples.}
\label{fig:cmnist:samples}
\end{center}
\end{figure*}

We show samples obtained on CelebA, CIFAR10 and ColorMNIST in Figures~\ref{fig:celeba:samples},~\ref{fig:cifar:samples} and~\ref{fig:cmnist:samples}, respectively.

\end{document}